%% file: main.tex
\documentclass[10pt,twocolumn,letterpaper]{article}

\usepackage{cvpr}      %

\input{preamble}
\definecolor{cvprblue}{rgb}{0.21,0.49,0.74}
\usepackage[pagebackref,breaklinks,colorlinks,allcolors=cvprblue]{hyperref}

\def\paperID{*****} %
\def\confName{CVPR}
\def\confYear{2025}
\def\ie{\emph{i.e}\onedot}

\newcommand{\xhdr}[1]{\vspace{2pt}\noindent\textbf{#1}}
\newcommand{\figref}[1]{Fig.~\ref{#1}}
\newcommand{\secref}[1]{Sec.~\ref{#1}}
\newcommand{\appref}[1]{App.~\ref{#1}}
\newcommand{\tabref}[1]{Tab.~\ref{#1}}
\makeatletter
\DeclareRobustCommand\onedot{%
  \futurelet\@let@token\@onedot
}
\def\@onedot{%
  \ifx\@let@token.\else.\null\fi\xspace
}
\makeatother
\usepackage{multirow} 
\usepackage{stfloats}
\usepackage{enumitem}

\usepackage{stfloats}
\newcommand{\algname}{Robots Imitating Generated Videos\xspace}
\newcommand{\algabrvname}{RIGVid\xspace}
\newcommand{\method}{\algname}
\newcommand{\methodshort}{\algabrvname}

\newcommand{\repr}{6D Object Pose Trajectory\xspace}

\newcommand{\reprlowercase}{6D pose rollout\xspace}
\usepackage{paralist}

\title{Robotic Manipulation by Imitating Generated Videos \\ Without Physical Demonstrations}

\author{
    Shivansh Patel$^{1}$ \quad
    Shraddhaa Mohan$^{1}$ \quad
    Hanlin Mai$^{1}$ \\
    Unnat Jain$^{2*}$ \quad
    Svetlana Lazebnik$^{1*}$ \quad
    Yunzhu Li$^{3}$\thanks{denotes equal advising.} \\
    \\
    $^{1}$UIUC \quad
    $^{2}$UC Irvine \quad
    $^{3}$Columbia University \\
    \\
    \url{https://rigvid-robot.github.io/}
}

\begin{document}
\maketitle
\input{sec/0_abstract}

\input{sec/1_intro}

\input{sec/2_related_works}
\input{sec/3_method}

\input{sec/4_experiments}

\input{sec/5_conclusion}
\input{sec/6_acknowledgement}
{
    \small
    \bibliographystyle{ieeenat_fullname}
    \bibliography{main}
}

\appendix
\input{sec/X_suppl}

\end{document}

%% file: sec/0_abstract.tex
\begin{abstract}
This work introduces \method (\methodshort), a system that enables robots to perform complex manipulation tasks—such as pouring, wiping, and mixing—purely by imitating AI-generated videos, without requiring any physical demonstrations or robot-specific training. Given a language command and an initial scene image, a video diffusion model generates potential demonstration videos, and a vision-language model (VLM) automatically filters out results that do not follow the command. A 6D pose tracker then extracts object trajectories from the video, and the trajectories are retargeted to the robot in an embodiment-agnostic fashion.
Through extensive real-world evaluations, we show that filtered generated videos are as effective as real demonstrations, and that performance improves with generation quality. We also show that relying on generated videos outperforms more compact alternatives such as keypoint prediction using VLMs, and that strong 6D pose tracking outperforms other ways to extract trajectories, such as dense feature point tracking. These findings suggest that videos produced by a state-of-the-art off-the-shelf model can offer an effective source of supervision for robotic manipulation.
\end{abstract}

%% file: sec/1_intro.tex
\section{Introduction}
\label{sec:intro}

Videos offer a rich and expressive source of training data for robotic manipulation, and numerous methods have successfully leveraged them for supervision. Such methods typically fall into two categories: (1) Learning from publicly available large-scale datasets of real-world videos~\cite{bahl2023affordances,ye2024latent,gao2025adaworld,bharadhwaj2023zero,chang2020semantic,sivakumar2022robotic}, and (2) Imitation of specific demonstrations captured under controlled conditions that closely match the execution setting~\cite{bahl2022human,kareer2024egomimic,wang2023mimicplay,chane2023learning,li2024okami,lepert2025phantom}. Unfortunately, both of these strategies come with challenges that limit broad deployment. Large-scale video datasets often introduce domain gaps~\cite{zhou2024mitigating,gao2025adaworld,xie2024decomposing} and require adaptation to specific robot embodiments and tasks~\cite{bahl2023affordances,o2023open}. On the other hand, video-based imitation involves laborious data collection that must ensure close alignment in viewpoints, morphologies, and interaction modalities~\cite{bahl2022human,dasari2021transformers,bahety2024screwmimic,sivakumar2022robotic}. %

\begin{figure*}[t]
  \centering
  \includegraphics[width=\textwidth]{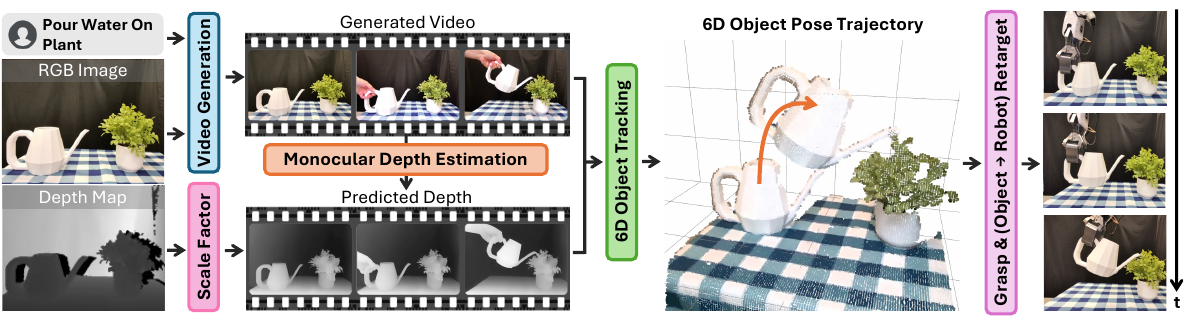}
  \vspace{-0.6cm}  %
  \caption{\textbf{\methodshort overview.}
  Given an initial scene image and depth, we generate a video conditioned on a language command. A VLM-based automatic filtering step (not shown) can be used to reject videos that fail to follow the prompt. A monocular depth estimator recovers depth for each frame of the generated video, and these depth maps are combined with the corresponding RGB frames to produce \repr. After grasping, the trajectory is retargeted to the robot for execution.}

  \label{fig:method}
  \vspace{-0.5cm}  %
\end{figure*}

Motivated by recent advances in video generation, we explore a potentially new paradigm: can a \textbf{single generated video}, synthesized to exactly match our input environment and task description, be used as the sole source of supervision for robotic manipulation? Recently released models like SORA~\cite{videoworldsimulators2024} and Kling~\cite{KlingAI2024} have demonstrated impressive capabilities in producing realistic-seeming videos from language and image inputs. 
At the same time, it has been shown that such videos can suffer from distorted object geometries~\cite{liu2024reconx,zhang2024world}, physically implausible interactions~\cite{motamed2025generative,yang2025vlipp}, and unrealistic scene dynamics~\cite{bansal2024videophy,guo2025t2vphysbench}. Consequently, while the idea of synthesizing video demonstrations is enticing, %
its usefulness in the robotics setting is yet to be convincingly established. Prior work incorporating video generation into robotics typically relies on additional supervision, such as task-specific training~\cite{du2024learning} or fine-tuning on offline robot trajectory datasets~\cite{bharadhwaj2024track2act,bharadhwaj2024gen2act}. By contrast, we ask whether a robot can perform real-world manipulation tasks solely by imitating generated videos---\textit{without any additional supervision or task-specific training}.

To this end, we introduce \textbf{\method (\methodshort)}, a framework that connects video generation models to real-world robotic execution. \figref{fig:method} gives an oveview of the method. Given an input RGB-D image of the scene and a free-form language command (e.g., ``pour water on the plant''), we use a state-of-the-art video diffusion model to generate a candidate video of the task being performed. The generated video is not guaranteed to accurately follow the language command -- but we show that a VLM can be used to automatically filter out unsuccessful generations with high accuracy. Next, we estimate per-frame depth on the video, segment the manipulated object, and track its \textit{6D object pose trajectory} across the frames using the FoundationPose tracker~\cite{wen2023foundationpose}. While this tracker relies on a pre-computed object mesh, preliminary experiments (\appref{sec:model_free_tracking}) indicate that our method is compatible with mesh-free approaches, though their inference speed is currently infeasible for real-time deployment. Finally, the extracted 6D object pose trajectory is retargeted to the robot for execution. The retargeting process only requires the transformation between the end-effector and the object, so it can be easily applied across platforms. %
During deployment, \methodshort performs real-time object tracking and dynamically adjusts the robot’s actions to handle disturbances and execution-time variations, promoting robust and adaptive behavior.

We evaluate \methodshort on four real-world manipulation tasks: pouring water, lifting a lid, placing a spatula on a pan, and sweeping trash. These tasks span diverse manipulation challenges, including a range of depth variation (minimal in pouring vs. significant in lifting), thin and partially occluded objects (spatula, sweeping brush), and diverse object geometries and actions. Our results show that, when paired with our filtering mechanism, generated videos are as effective as human videos for visual imitation, enabling robots to act entirely from synthetic supervision. Moreover, the performance of \methodshort improves with video quality, suggesting a favorable trend where advances in generative models directly translate to stronger manipulation capabilities. 

The main downside of video generation is its substantial computational cost. Also, on a representational level, one may wonder whether predicting video pixels is wasteful, and whether we should instead predict a more compact and minimal representation that can be efficiently translated to an executable trajectory. One example of this philosophy is the recent ReKep method~\cite{huang2024rekep}, which uses a VLM to generate relational keypoint constraints from a task description and then solves for a 6D trajectory given these constraints. We compare our approach to ReKep and demonstrate that video generation does, in fact, perform substantially better than the generation of a more sparse and high-level representation. 
Next, given a generated video, one may ask whether 6D object-level tracking is necessary, given its up-front requirement of an object mesh. To address this question, we compare against a broad range of alternative tracking approaches --- sparse point tracking~\cite{bharadhwaj2024track2act}, dense optical flow~\cite{ko2023learning}, 3D feature fields~\cite{kerr2024robot}, and generated goal supervision~\cite{bharadhwaj2024gen2act} --- and show consistently higher success rates. %

In summary, our key contributions are:
(1) We propose a framework that enables robots to perform open-world manipulation tasks without any real-world demonstrations --- only generated videos. %
(2) We show high-quality generated videos perform on par with real videos for robotic imitation, establishing that synthetic data can serve as an effective substitute for real data in this domain. %
(3) We demonstrate that our combination of video generation and 6D trajectory extraction outperforms a wide variety of competing state-of-the-art methods based on VLMs, point tracking, optical flow, feature fields, and generated-goal supervision.

%% file: sec/2_related_works.tex
\section{Related Work}
\noindent\textbf{Direct Imitation from Videos.} This seeks to match visual states in demonstration videos to those of the learner, without requiring expert action labels or robot state information~\cite{dasari2021transformers,valassakis2022demonstrate,pathak2018zero,bahl2022human,sharma2019third,yu2018one,kareer2024egomimic,sivakumar2022robotic,fu2024humanplus,wang2023mimicplay,shi2025zeromimic,kerr2024robot,hsu2024spot,xu2023xskill,wang2024one}. While effective, this approach demands paired demonstrations in the same setting. 
A common strategy is to leverage visual correspondences—tracks~\cite{bharadhwaj2024track2act} or optical flow~\cite{argus2020flowcontrol,xu2024flow,gao2024flip}—to infer object trajectories. Bharadhwaj \etal~\cite{bharadhwaj2024track2act} predicts object tracks and uses PnP to recover poses for closed‑loop task execution. Dense descriptor learning~\cite{florence2018dense,zhu2024densematcher,vecerik2024robotap} has proven powerful for handling variations in object geometry and appearance. Kerr \etal~\cite{kerr2024robot} recover object part trajectories from monocular videos using feature fields. 
Crucially, these methods rely on demonstrations collected under conditions closely matching the target task. In contrast, our method removes this requirement by generating task and scene-conditioned videos.

\noindent\textbf{Imitation from Offline Videos.} This paradigm alleviates the need for paired demonstrations by leveraging offline video data, and has consequently attracted significant attention~\cite{smith2019avid,liu2018imitation,sharma2018multiple,zakka2022xirl,sermanet2018time,Finn2017OneShotVI,vlamp,mandlekar2018roboturk,ko2023learning,barcellona2024dream,zhou2025you,wang2023mimicplay,chang2020semantic,ponimatkin20256d,patel2022learning,ren2025motion,shi2025zeromimic}. Many works focus on learning affordance models from internet‑scale video datasets~\cite{bahl2023affordances,srirama2024hrp,mendonca2023structured,li2024one,yuan2024general,ju2024robo,li2024learning,dasari2023unbiased,karamcheti2023language,baker2022video,chang2020semantic,ponimatkin20256d,patel2022learning,ren2025motion}. Here, affordances are defined as scene-centric predictions of where and how an agent can interact, often captured as contact-point heatmaps and short motion trajectories that can be translated into robot actions. For example, Bahl~\etal~\cite{bahl2023affordances} learn from large-scale human videos to output dense contact maps and trajectory waypoints, which downstream imitation, exploration, or reinforcement modules can transform into executable robot motions. However, these methods suffer from domain gap between training videos and task‑specific environments, and require additional mechanisms to obtain task-conditioned affordances. In contrast, our method does not explicitly predict affordances, but instead relies on generated, task- and scene-specific generated videos for imitation.

\noindent\textbf{Video Generation for Robotics.} Video generation has emerged as a promising avenue for robotics~\cite{du2024learning,du2023video,ajay2023compositional,liang2024dreamitate,bharadhwaj2024gen2act,liang2024dreamitate,zhen2025tesseract,albaba2025nil,yang2023learning}. A common limitation of these is their reliance on robot data, either to train the video generation model~\cite{liang2024dreamitate,sun2024video}, or to train policies~\cite{bharadhwaj2024gen2act}, or both~\cite{du2024learning,du2023video,ajay2023compositional}. Bharadhwaj \etal\cite{bharadhwaj2024gen2act} leverages tracks on generated videos to condition policy learning. Albaba \etal~\cite{albaba2025nil} uses generated videos to compute rewards for RL training. The most closely related work is Liang \etal~\cite{liang2024dreamitate}, which executes robotic tasks by tracking tools attached to the robot’s end effector. While effective, their method relies on 1,822 human-collected robot demonstrations for four tasks, and is confined to tasks executable only by tools. In contrast, our approach requires no such data collection. Instead of tools, our method tracks objects, enabling a significantly broader range of manipulation tasks without using any robot data.

\noindent\textbf{6D Pose Estimation and Tracking.}
Instance-level object pose tracking methods fall into two main categories: model-based and model-free. Model-based approaches~\cite{he2020pvn3d,he2021ffb6d,labbe2020cosypose,park2019pix2pose,labbe2022megapose,shugurov2022osop,nguyen2024gigapose, caraffa2024freeze} require a 3D CAD model and typically estimate pose by constructing 2D-3D correspondences and solving the P$n$P problem~\cite{park2019pix2pose,sling,xgx,tremblay2018deep,lepetit2009ep}. In contrast, model-free methods~\cite{he2022onepose++,he2022fs6d,park2020latentfusion,sun2022onepose,cai2020reconstruct,li2023nerf,liu2022gen6d} rely on multiple reference images instead of an explicit 3D mesh. Recent advances in vision foundation models and large datasets have enabled zero-shot methods~\cite{ausserlechner2024zs6d, labbe2022megapose, ornek2024foundpose, caraffa2024freeze, liu2024unopose}, which extend to unseen objects and categories but still lag behind instance-level methods in performance. We employ FoundationPose~\cite{wen2023foundationpose}, a versatile instance-level tracking method that supports model-based pose tracking. Notably, it does not require any instance-specific fine-tuning. Our choice is guided by its state-of-the-art performance and real-time execution speed, both of which are crucial for ensuring robustness against disturbances during execution.

\noindent\textbf{Motion Retargeting for Object Manipulation.} Early work in learning from demonstration established the foundation for object-centric motion retargeting~\cite{gleicher1998retargetting, calinon2016tutorial, niekum2012learning, luo2023perpetual, peng2021amp, jiang2023motiongpt}. More recently, deep learning-based retargeting methods have emerged~\cite{cheng2024expressive, choi2020nonparametric, he2024learning}, with some incorporating object-centric representations to bridge the gap between the demonstrator and the robot~\cite{kerr2024robot, li2024okami, wu2024one}. Many approaches have applied retargeting to humanoid robots~\cite{nakaoka2005task, hu2014online, penco2019multimode, kuindersma2016optimization, liang2021dynamic}. Other works have extended these techniques to dexterous manipulation~\cite{qin2022dexmv, lakshmipathy2024kinematic}. Like most prior work, we assume a fixed transformation between the end-effector and the object. While motion retargeting has traditionally relied on human demonstrations, \algabrvname eliminates this dependency by leveraging generated videos.

%% file: sec/3_method.tex
\section{Our Method: \method}
\label{sec:method}
An overview of our method is shown in \figref{fig:method}. It takes as inputs the initial scene RGB image, its corresponding depth map, and a free-form human command. Our goal is to predict the robot's 6DoF end-effector trajectory. %
This section describes the key steps of \methodshort: (1)~Generate a scene and task-conditioned video and predict its corresponding depth using a monocular depth estimator (\secref{sec:generating_vid}); (2)~Compute \reprlowercase via an object pose tracker (\secref{sec:objectTracking}); (3)~Grasp the object and retarget the pose trajectory to the robot, and execute the resulting trajectory (\secref{sec:robotTracking}).

\subsection{Generating Videos and Corresponding Depth}
\label{sec:generating_vid}

Since the generated videos may not necessarily follow the language command or have other issues, we need an automatic filtering mechanism to discard inaccurate generations. We found that we can do the filtering reliably by prompting a VLM -- specifically, GPT-4o~\cite{achiam2023gpt} -- to assess whether the generated video depicts a successful execution of the command. As image input to GPT-4o, we sample four evenly spaced frames in the video and concatenate them vertically to create a video summary. The VLM determines whether the action described in the command is performed by a visible hand. \appref{sec:appendix_prompting_for_filtering} provides the full prompt used for filtering and examples of video summaries with their corresponding VLM responses. If the response is negative, we regenerate the video and repeat the process for up to five attempts. If all attempts fail, we default to the final attempt. 

As input to the downstream tracking step, we also need to predict the depth for the generated video, using the predictor from Ke \etal~\cite{ke2024rollingdepth}. One complication is that the estimated depth values are not grounded in real-world units, but subject to a scale and shift ambiguity~\cite{hartley2003multiple}. Consistent with prior works adopting depth estimators in vision-based robotics~\cite{gervet2023navigating,chang2023goatthing}, we compute an affine scale-and-shift transformation, aligning the predicted depth in the first frame with the initial real depth map around the active object (discussed in \secref{sec:objectTracking}). This transformation is then applied to the entire predicted video to resolve the ambiguity.

\subsection{Identifying Active Object Mask and \repr} \label{sec:objectTracking}
To extract \reprlowercase, we first identify the active object—the one being manipulated in the generated video. A binary mask for this object in the initial RGB image is essential for object tracking and determining which object to grasp. Given the initial image and the task command, we prompt GPT-4o to identify the object most likely to be manipulated. We then ground the predicted object category into a bounding box using Grounding DINO~\cite{liu2023grounding}, and further refine this into a segmentation mask using SAM-2~\cite{ravi2024sam2segmentimages}.

Once the active object is localized by the mask, we track it across the generated video using the scaled predicted depth. This yields the \reprlowercase. Tracking objects in videos is a rich area of research, and we experimented with several 6D pose trackers~\cite{labbe2022megapose,wen2023bundlesdf,wen2023foundationpose}. For real-world deployment, we found FoundationPose~\cite{wen2023foundationpose} to perform the best. It requires an object mesh, which we pre-compute using BundleSDF~\cite{wen2023bundlesdf}. For this, we record a short RGBD video where the object is held and rotated in front of the camera to capture all sides. While straightforward, this process constrains our method to settings where a mesh can be precomputed. Nonetheless, as shown in \appref{sec:model_free_tracking}, our method is also compatible with mesh-free approaches—BundleSDF can jointly reconstruct and track the object—but current inference speeds make these alternatives infeasible for real-time use. To ensure stable and realistic motion during execution, we apply an averaging filter to smooth abrupt pose changes, particularly in rotation. Additional details on this smoothing step are provided in \appref{sec:remove_traj_noise}.

\subsection{Object to Robot Motion Retargeting} \label{sec:robotTracking}

\begin{figure}[h]
    \centering
    \includegraphics[width=\linewidth]{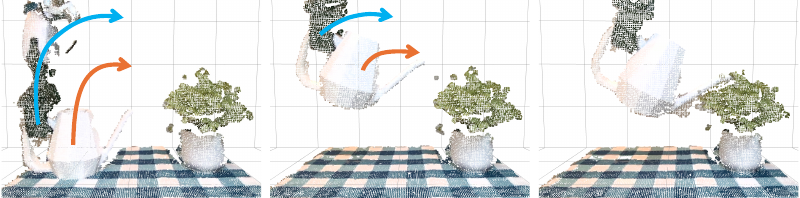}
    \caption{\textbf{Re-targeting \algabrvname to a robot trajectory.} Assuming a fixed transformation between the end-effector and the object after grasping, the \repr (\textit{orange arrow}) is re-targeted to the robot (\textit{blue arrow}). This formulation is embodiment agnostic and can be transferred to a different robot.}
    \label{fig:retargeting}
\end{figure}

Once the object trajectory is obtained, we first grasp the object. We use an off-the-shelf grasper, AnyGrasp~\cite{fang2023anygrasp}, to identify and execute the highest-scoring grasp within a defined boundary around the active object mask. After grasping, we retarget its trajectory to the robot's end-effector. Since the object remains firmly grasped, we assume a fixed transformation between the robot’s end-effector and the object. This transformation is obtained by composing two rigid-body transforms: (1) the pose of the object relative to the gripper at the moment it is grasped and (2) the offset between the gripper and the robot's end-effector. By combining these two components, we obtain a single transformation from the end-effector to the object.

The corresponding end-effector trajectory is obtained by applying the fixed end-effector-to-object transformation to the object’s pose along the entire trajectory. This formulation ensures that the retargeted \reprlowercase follows the object's motion while maintaining a stable grasp. These are executed on the physical robot, enabling it to reproduce the object's movement as observed in the generated video. A key strength of this approach is that it is robot-agnostic. Specifically, to accommodate a different robot or gripper, only the end-effector to the object transformation needs to be updated to reflect the new end-effector configuration.
\begin{figure}[h!]
\centering
\includegraphics[width=\linewidth]{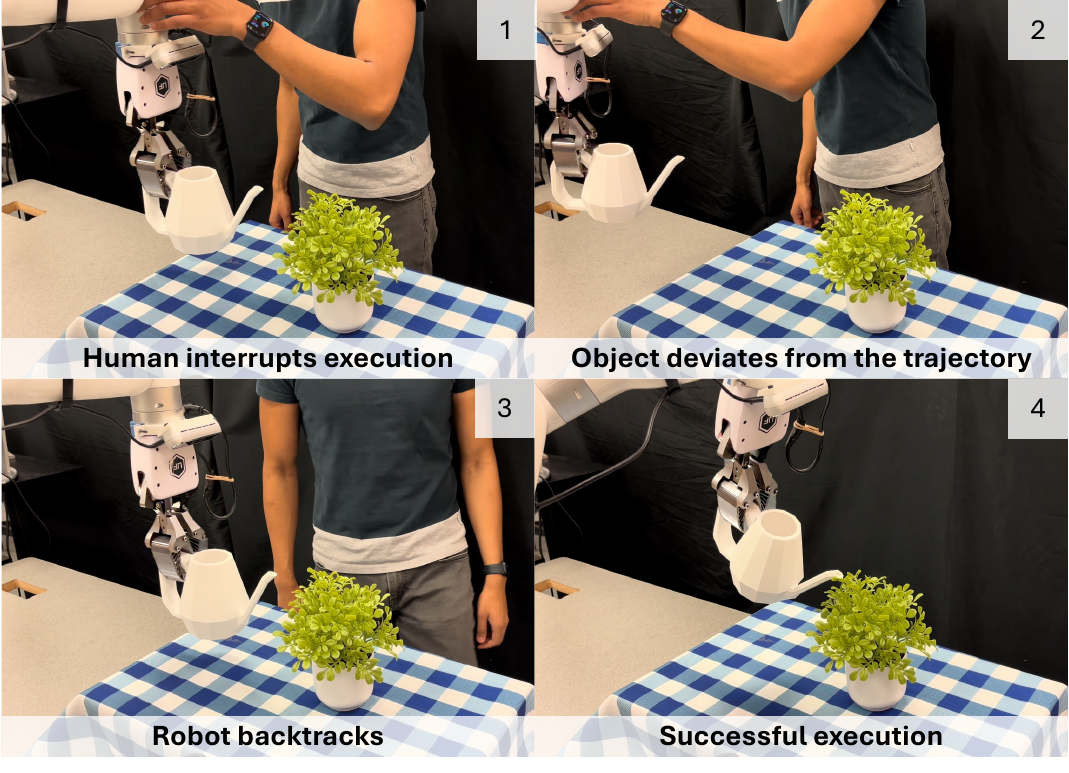}
\vspace{-0.6cm}
\captionsetup{type=figure}
\caption{\small{\textbf{\methodshort is robust to perturbations.} A human pushes the robot during execution (image 1), causing the object to deviate from the planned trajectory. When the deviation is detected (image 2), the robot backtracks to the last successfully executed trajectory point (image 3) and then resumes the planned motion (image 4).}}
\label{fig:robustness}
\end{figure}

\begin{figure*}[t]
  \centering
  \begin{minipage}{\linewidth}
    \centering
    \includegraphics[width=\linewidth]{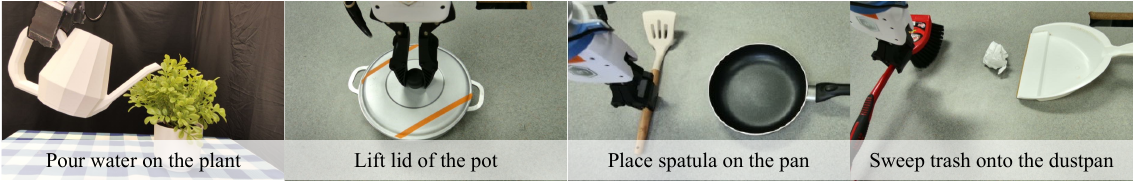}
  \end{minipage}
  \captionsetup{type=figure}
  \vspace{-8pt}
  \caption{\small{\textbf{Evaluation tasks.} We evaluate \methodshort on everyday manipulation tasks of varying difficulty.}}
  \label{fig:tasks}
\end{figure*}

\subsection{Closed Loop Execution}

A core strength of our approach is its ability to operate in a closed-loop manner, enabling robust execution despite disturbances or unexpected changes during task execution. During deployment, the system continuously tracks the object’s 6D pose in real time using FoundationPose to update the robot’s end-effector trajectory as the task progresses. This feedback allows the robot to dynamically adjust its motions: if the object deviates from the planned trajectory due to external perturbations, such as a human pushing the robot or a slip after grasping, the system detects the discrepancy by comparing the current object pose to the precomputed trajectory. If the detected deviation exceeds a threshold of 3 cm in position or 20 degrees in orientation, the robot backtracks to the last successfully executed trajectory point and resumes execution from there (\figref{fig:robustness}). This recovery mechanism enables \methodshort to maintain stable task execution, realigning and successfully completing the manipulation task despite perturbations. Additional examples of robustness to perturbations are provided in \appref{sec:more_robustness}.

%% file: sec/4_experiments.tex
\section{Experiments}
\label{sec:experiments}
This section presents our experimental evaluation. We describe the robot setup, manipulation tasks, and evaluation protocol (\secref{sec:setup}). Then assess the impact of video generation models and filtering strategies on downstream robotic performance (\secref{sec:video_model_comparison}). Next, we compare \algabrvname to SOTA VLM-based trajectory prediction method that allows zero-shot execution (\secref{sec:ours_vs_vlm}), and to alternative tracking approaches for trajectory extraction (\secref{sec:baselines}). Finally, we test generalization across embodiments, extensions to new tasks, and robustness to real-world disturbances (\secref{sec:robust_error_analysis}).

\subsection{Robot Setup, Tasks, and Evaluation}
\label{sec:setup}
We conduct experiments on an xArm7 robot arm with a stationary Orbbec Femto Bolt camera, positioned next to the robot to capture RGBD observations. We evaluate our method on four everyday manipulation tasks, which are illustrated in \figref{fig:tasks}. These span a diverse range of robotic challenges, and their descriptions are as follows:
\begin{compactenum}
\item \textbf{Pouring water} requires the robot to position and tilt a watering can above a plant. While the depth of the can relative to the camera remains largely constant, successful execution demands a smooth trajectory spanning the pick-up, transport, and pouring phases. A trial is considered successful if the watering can's spout is positioned above the plant at the end of the execution.

\item \textbf{Lifting a lid} requires the robot to lift a pot lid. Unlike pouring, where the camera is viewing the scene from the side, the camera here is looking down towards the pot. As a result, this task involves significant variation in object depth, as the lid moves closer to the camera during execution. %
Success is achieved if the lid is no longer in contact with the pot at the end of the trial.

\item \textbf{Placing a spatula on a pan} requires the robot to place the head of a spatula into a pan. The spatula has a thin, elongated geometry and is often partially occluded during manipulation, which presents a challenge for object tracking, particularly for non-mesh-based approaches. The task is considered successful if the spatula’s head is in the pan at the end of execution.

\item \textbf{Sweeping trash} requires the robot to sweep trash into a dustpan. This task is especially challenging as it combines the need for precise positioning to align the head of the sweeping brush with the trash, along with all the challenges encountered from the placing task. A trial is successful if the trash is touching the base of the dustpan at the end of the execution. 

\end{compactenum}

Task success is determined via human judgment based on the above criteria, though the procedure could be readily automated with a VLM.
The initial setup configuration is fixed across trials of the same task,  and each trial has a different generated video. 
All baselines use the same videos for consistent comparison and reporting.

\subsection{Quality and Filtering of Generated Videos}
\label{sec:video_model_comparison}

\begin{figure*}[t]
    \centering
     \includegraphics[width=\linewidth]{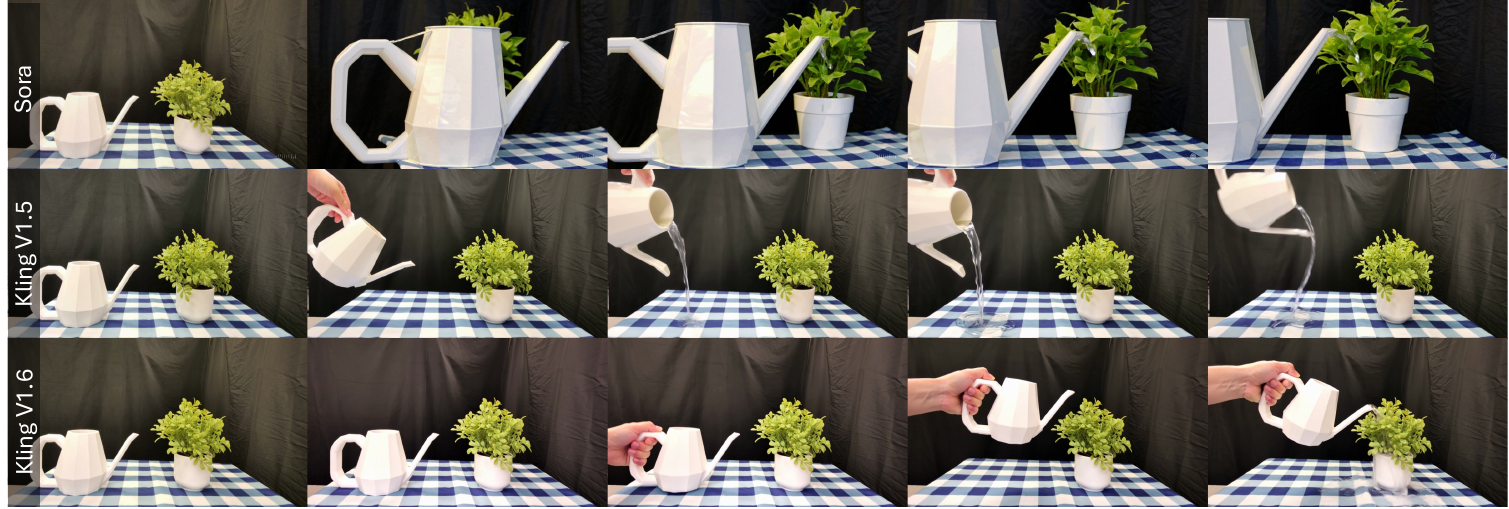}
    \captionsetup{type=figure}
    \vspace{-15pt}
    \caption{\small{\textbf{Qualitative comparison of video generation for three models.} Sora (top) drastically alters the scene layout and object size. Kling v1.5 (middle) does not fully follow the prompt (water not poured over the plant) and exhibits physically implausible behaviors (water pouring out of the top of the kettle but not the spout). Kling v1.6 (bottom) produces the most consistent and realistic result.}}
    \label{fig:main_qualitative_model_compare}
    \vspace{-10pt}
\end{figure*}

As discussed in \secref{sec:generating_vid}, we experimented with Sora, Kling v1.5, and Kling v1.6 for video generation. We also compared different video filtering strategies. Next, we present our key empirical findings.

\textit{How do different video generation models compare for robotic imitation?} Sora is known for creating visually impressive and cinematic videos. Unfortunately, these videos often prioritize aesthetics over following the human command. As seen in the top row of \figref{fig:main_qualitative_model_compare}, Sora frequently alters the camera viewpoint, changes object positions, or even swaps out objects mid-sequence. This lack of scene and object consistency makes Sora poorly suited for imitation. Kling v1.5 places more emphasis on following language instructions, generally preserves the original scene, and correctly depicts the target object. Nonetheless, it is still prone to physically implausible behaviors and command following failures. In the second row of \figref{fig:main_qualitative_model_compare}, the teapot is not positioned over the plant and the water pours out from the top, not the spout (in other failure cases, nothing at all happens in the video, and the command is not even attempted). %
By contrast, Kling v1.6 (bottom row of \figref{fig:main_qualitative_model_compare}) has greatly improved command following and physical plausibility, proving to be the most reliable video generator for us. More examples of generated videos are shown in App. \figref{fig:large_qualitative_model_compare}.

\begin{figure}[h]
    \centering
     \includegraphics[width=\linewidth]{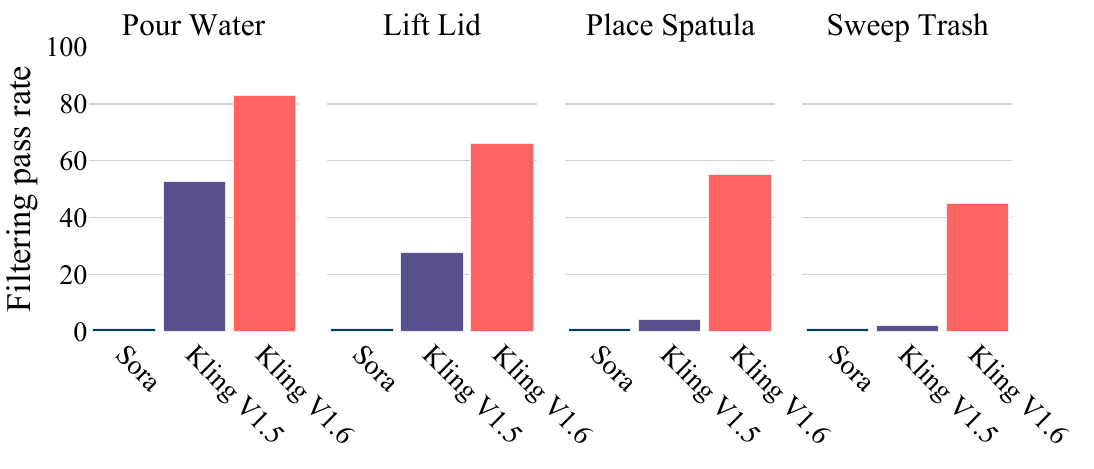}
     \vspace{-0.6cm}
    \captionsetup{type=figure}
    \caption{\small{\textbf{Filtering statistics.} Kling V1.6 videos have the highest pass rate, demonstrating more accurate adherence to language commands.}}

    \label{fig:filtering_stats}
    
\end{figure}

\textit{What are the filtering statistics for different video generation models?} 
Confirming the trends described above, \figref{fig:filtering_stats} reports the pass rates of each model across our four tasks for the GPT-4o filter described in \secref{sec:generating_vid}. Sora fails all tasks 100\% of the time. Kling v1.5 does better, successfully passing pouring 52.6\% of the time, lifting 27.7\%, placing 4\%, and sweeping 2\%. Kling V1.6 shows a substantial improvement across tasks, passing pouring 83\%, lifting 66\%, placing 55\%, and sweeping 45\% of the time. We noticed that, particularly for the placing and sweeping tasks, even Kling V1.6 frequently generated videos in which the command was not followed. In many cases, the video remained static, and no action was performed.

\begin{table*}[t]
    \centering
    \resizebox{0.7\textwidth}{!}{
    \begin{tabular}{lccccc}
        \toprule
        \textbf{Filtering Metrics} & \textbf{Pour Water} & \textbf{Lift Lid} & \textbf{Place Spatula} & \textbf{Sweep Trash} & \textbf{Average} \\
        \midrule
        \textbf{Video-text Consistency} &  0.06 & 0.47 & 0.70 & 0.11 & 0.34 \\
        \textbf{I2V Subject Consistency} &  0.35 & 0.58 & -0.09 & 0.63 & 0.37 \\
        \textbf{Querying GPT o1} &  \textbf{0.91} & \textbf{0.91} & \textbf{0.91} & \textbf{0.66} & \textbf{0.84} \\
        \bottomrule
    \end{tabular}
    }
    \caption{\small{\textbf{Comparison of video filtering metrics.} Pearson correlation coefficients measure each metric’s effectiveness in assessing whether
a generated video follows the language command. Querying GPT o1 proves to be most effective.}}
    \label{tab:compare_video_metrics}
\end{table*}

\textit{How accurate is VLM-based video filtering, and are there any simpler alternatives?} In \tabref{tab:compare_video_metrics}, we report Pearson correlation coefficients between filtering metrics and human judgments of generation correctness. Our VLM-based filtering achieves strong agreement with human ratings across all tasks, with high correlation values. Most errors made by the VLM-based filter are false negatives—occasionally discarding usable videos, but almost never passing an incorrect one. We also explore the most relevant metrics for our case from a recent benchmark suite for evaluating video generation quality and instruction following, VBench++~\cite{huang2024vbench++}: (i) video-text consistency measuring the alignment between the command and the generated video~\cite{wang2023internvid}, and (ii) image-to-video (I2V) subject consistency which evaluates whether subjects present in the input image persist throughout the video~\cite{caron2021emerging}. These metrics correlate only weakly or inconsistently with task success and are not reliable for filtering. 

\begin{figure*}[t]
  \centering
  \begin{minipage}{\linewidth}
    \centering
    \includegraphics[width=0.8\linewidth]{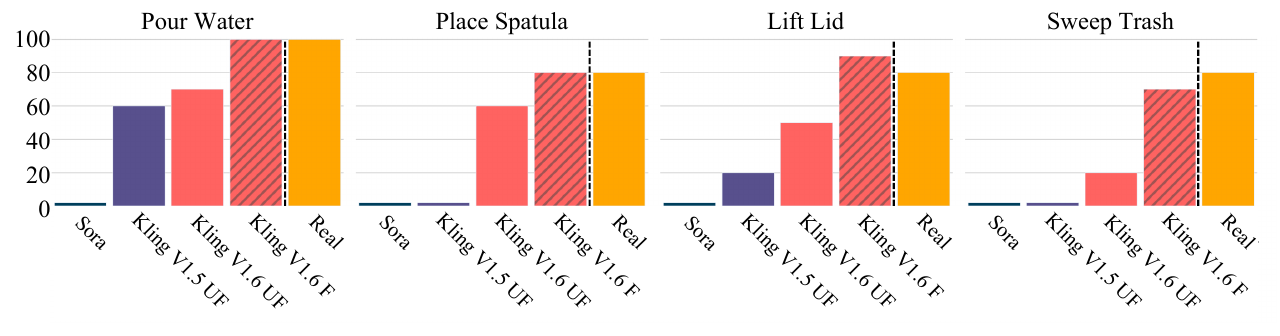}
  \end{minipage}
  \vspace{-0.4cm}
  \captionsetup{type=figure}
  \caption{\small{\textbf{\algabrvname performance vs.\ video quality.} The dashed lines separate performance on generated videos from real videos. Kling V1.6 produces most reliable videos and leads to highest \methodshort success. Filtered videos perform on par with real ones. UF denotes unfiltered and F denotes filtered. }}
  \label{fig:video_gen_results_bar}
  \vspace{-10pt}
\end{figure*}

\textit{Does higher video quality lead to better robot performance?} To quantify this, \figref{fig:video_gen_results_bar} plots \algabrvname’s task success across five video sources: unfiltered Sora, unfiltered Kling v1.5, unfiltered Kling v1.6, filtered Kling v1.6, and real human demonstration videos. For each source, we use 10 videos per task. We observe a clear trend: as video quality improves, so does success rate. Sora’s unfiltered videos lead to 0\% success rate, Kling v1.5 performs better, and Kling v1.6 gives the highest results among all generated videos. Filtering dramatically improves reliability: after discarding failed generations using our automatic approach, success rate with filtered Kling v1.6 videos improves from 80\% to 100\% on pouring, from 60\% to 80\% on lifting, from 50\% to 90\% on placing, and from 20\% to 70\% on sweeping.

\textit{Can generated videos replace real videos for imitation?} The results in \figref{fig:video_gen_results_bar} indicate that, when using filtered Kling v1.6 videos, \methodshort’s performance is similar to that achieved with real human demonstration videos. This finding suggests that, at current model quality, generated videos are already sufficient for visual imitation, substantially reducing the need for manual data collection.

\textit{What causes failure of imitation given high-quality videos?} Aside from one case where the object slipped out of the gripper, all failures on filtered Kling v1.6 videos are attributed to errors in monocular depth estimation. These errors result in inaccurate 6D trajectories and lead to tracking failures. Notably, similar depth estimation issues are also observed in real videos, suggesting that the limitation lies in the depth model itself. \appref{sec:depth_estimation_errors} provides a detailed analysis of failure cases with qualitative examples. 

\begin{figure}[b!]
    \centering
     \includegraphics[width=\linewidth]{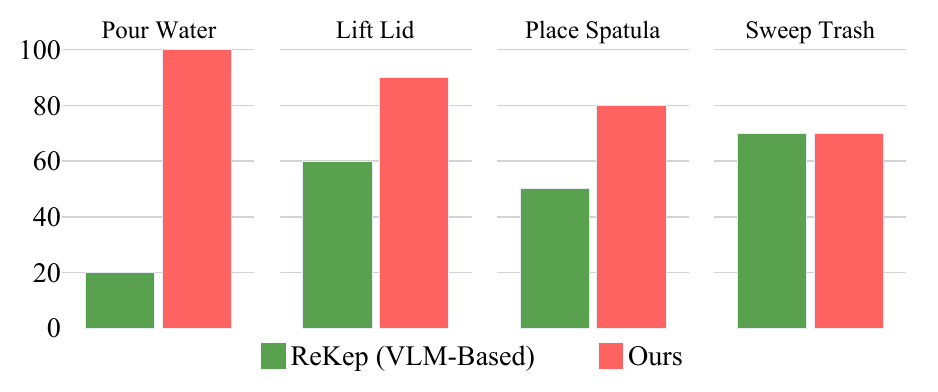}
    \captionsetup{type=figure}
    \vspace{-0.7cm}
    \caption{\small{\textbf{\methodshort vs. ReKep Success Rates.} \methodshort  outperforms SOTA VLM-based trajectory prediction method ReKep.}}
    \label{fig:ours_vs_llm}
    \vspace{-0.1cm}
\end{figure}

\subsection{\methodshort vs. VLM-Based Trajectory Prediction}
\label{sec:ours_vs_vlm}

Video generation is computationally expensive, prompting the question of whether more efficient alternatives can enable robot manipulation without any demonstrations. VLMs offer one potential alternative by predicting simplified task abstractions—goal states~\cite{huang2023voxposer}, constraints~\cite{huang2024rekep}, or reward functions~\cite{patel2025real}—without generating full visual sequences, making them cheaper in computation and inference time. We take the state-of-the-art ReKep~\cite{huang2024rekep} method as a representative of this line of work, and compare against it in \figref{fig:ours_vs_llm}. In our comparison, \methodshort achieves 85\% vs. ReKep's 50\% success over four tasks. \appref{sec:rekep_appendix} illustrates ReKep’s failures, which we attribute to inaccurate keypoint predictions. This comparison suggests that, for our tasks and experimental setup, VLM-generated abstractions are compact and may lack the rich, necessary details for successful robot execution. Thus, despite its higher cost, video generation provides crucial supervision rather than being a wasteful expense in these settings.

While this result highlights, for our tasks and setup, the additional detail in generated videos supports more reliable execution than the current VLM-based alternative, it does not rule out the possibility that future or alternative VLM-based approaches could close this gap. Our findings suggest that, at present, video generation can provide richer supervision for manipulation compared to this specific VLM-based method, despite its higher computational cost.

\begin{table*}[t]
    \centering
    \resizebox{\textwidth}{!}{
        \begin{tabular}{p{1.6cm}p{7cm}p{2.6cm}p{8cm}}
            \toprule
            Method & Inputs & Intermediate Repr. & Salient Method Characteristic \\
            \midrule
            Track2Act & Initial RGBD, goal image & 2D point tracks & Only needs initial and goal image; no intermediate frames \\
            AVDC & Initial RGBD, task desc., mask, generated video & Optical flow & Dense flow over full video for trajectory optimization \\
            4D-DPM & 3D Gaussian field, generated video & 3D feature field & 3D field tracking (NeRF-like); needs 360$^\circ$ video \\
            Gen2Act & Initial RGBD, task desc., mask, generated video & 2D point tracks & Sparse tracks from generated video for pose estimation \\
            \bottomrule
        \end{tabular}
    }
    \vspace{-2pt}
    \caption{\small{\textbf{Summary of trajectory extraction baselines.} Each baseline processes the same generated videos to extract object trajectories for robot execution, but differs in inputs, intermediate representations, and the way correspondences are established.}}
    \label{tab:traj_baselines}
\end{table*}

\subsection{Comparison to Alternative Trajectory Extraction Methods}
\label{sec:baselines}
\begin{figure*}[t]
    \centering
        \centering
        \includegraphics[width=0.85\linewidth]{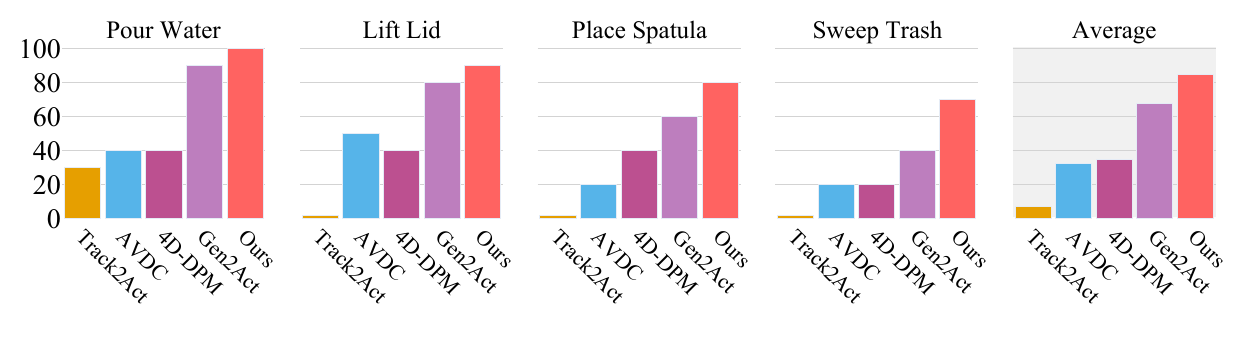}
    \captionsetup{type=figure}
    \vspace{-0.6cm}
    \caption{\small{\textbf{Comparative evaluation of trajectory extraction methods.} \algabrvname consistently achieves higher success rates across all four tasks; relative improvements are higher as tasks become harder (\ie, from left to right). }}
    \vspace{-0.1cm}
    \label{fig:main_results_bar}
\end{figure*}

Next, we investigate the best way to extract trajectory information from videos for the purpose of visual imitation. To this end, we adapted several competitive methods that use different types of tracking tos imitate a video without any demonstrations. Summary of all the methods is shown in \tabref{tab:traj_baselines}. For each method, we describe its inputs and outputs, core approach, our modifications, and the motivation for its inclusion (additional details can be found in \appref{sec:append_baselines}).  

\noindent\textbf{Track2Act~\cite{bharadhwaj2024track2act} (Tracks-Based)}. This method takes an RGBD image of the initial scene, and a single goal image that specifies the desired final configuration. Since we have no other way to get the goal image, we set it to the last frame of the generated video. Using only this pair of images, Track2Act uses a learned model to predict a dense grid of 2D point tracks, producing pixel-level correspondences between the initial and goal image. These tracks are then lifted to 3D using the depth map from the initial frame and converted into a sequence of 3D object poses via the Perspective-n-Point (PnP) algorithm. We do not finetune their track prediction network, and do not use their residual policy. Track2Act is an attractive alternative as it uses a dedicated track prediction network that operates solely on the start and goal images, without requiring any intermediate frames. However, its main drawback is that the learned track prediction network may not generalize to all scenarios, as evidenced by our experiments and qualitative results.

\noindent\textbf{AVDC~\cite{ko2023learning} (Flow-Based)}. Given an initial RGBD image, task description, and active object mask, AVDC predicts object motion by first generating a task-conditioned video and then computing optical flow between frames. This optical flow is used in an optimization process to reconstruct the object trajectory. In our adaptation, we substitute AVDC’s original video generator with our improved model, while preserving all downstream processing. Unlike Track2Act, AVDC leverages optical flow across the entire video, giving it dense temporal correspondences at every pixel and thus many more cues for tracking. It is attractive because it offers a denser input for object tracking. Nevertheless, it is sensitive to cumulative errors in flow estimation, which can degrade the accuracy of the resulting object trajectories.

\noindent\textbf{4D-DPM~\cite{kerr2024robot} (Feature Field-Based)}. This method takes a 3D Gaussian splatting field of the object and a real video of the task, and outputs object trajectories over time. A feature field, similar to NeRF representations, is a continuous mapping from 3D space to high-dimensional feature vectors that capture both geometry and appearance. By aligning the feature field with individual video frames, the method can estimate object trajectory across the video. To build the field, 4D-DPM requires a separate video where the object is captured from all sides. In our adaptation, since 4D-DPM expects a real human demonstration video, we instead use a generated video as the task input video. We modify this method from tracking object part poses to tracking single objects. This approach is compelling because it applies semantic, feature-based reasoning to track objects, capturing entire object structure from video, without relying on explicit correspondences. However, it tends to produce unstable tracking in our experiments, limiting its practicality. 

\noindent\textbf{Gen2Act~\cite{bharadhwaj2024gen2act} (Generated Goal-Based)}. Gen2Act takes as input an RGBD image of the scene and a task description, and outputs robot actions predicted by a learned policy. In the original formulation, the extracted tracks on the generated video were used to supervise behaviour-cloning on a large offline robotics dataset. In our adaptation, we do not use any policy learning. Instead, we extract object tracks from the generated video and directly estimate object poses from these tracks using the initial depth image. This removes any dependence on expert demonstration data or learned policies. Gen2Act is notable for leveraging sparse correspondences extracted from the generated video, enabling task-relevant object motion to be tracked and retargeted without requiring explicit actions. However, when large portions of the object become occluded or undergo significant rotations, many of the tracking points are lost, resulting in too few correspondences to estimate object pose accurately and ultimately causing the tracking to fail.

\begin{figure*}[!b]
    \centering
    \includegraphics[width=\linewidth]{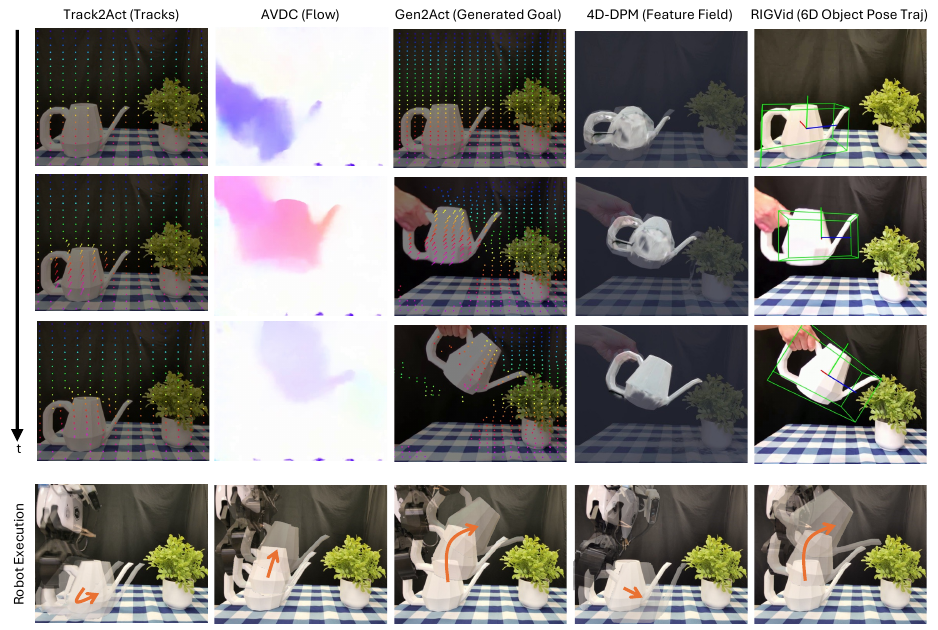}
    \captionsetup{type=figure}
    \vspace{-15pt}
    \caption{\textbf{Analyzing intermediate visual representations.} Only Gen2Act and our \repr can correctly track the position and rotation of the watering can, leading to a successful execution. Check the description in the main paper for detailed discussions of the failure modes of the alternative methods.}
    \label{fig:placeholder}
    \vspace{-12pt}
\end{figure*}

\figref{fig:main_results_bar} shows that RIGVid achieves a success rate of 85.0\%, compared to 67.5\% for Gen2Act and considerably lower rates for all other baselines. This margin grows with more complex tasks. Methods such as Track2Act (7.5\%), AVDC (32.5\%), and 4D-DPM (35.0\%) rely on point tracks or optical flow, but their performance remains limited—especially as object rotations or occlusions are severe. Gen2Act, which combines video generation with point-based tracking, closes part of the gap but consistently struggles when large portions of the object become untrackable. In contrast, RIGVid’s use of a structured 6D object pose trajectory enables robust execution across all tasks, accounting for the 17.5\% improvement over Gen2Act. This advantage persists when more powerful tracking models like CoTracker3~\cite{karaev2024cotracker3} are used, as shown in \appref{sec:gen2act_cotracker}. %

Looking at the task-wise breakdown in \figref{fig:main_results_bar}, RIGVid maintains high success rates even as object depth varies significantly (such as in the lifting task) or when the objects are thin, small, or partially occluded (such as in placing a spatula or sweeping trash). Other methods frequently struggle in these settings, often failing to recover accurate object trajectories when objects become partially hidden or change distance rapidly. The advantage of RIGVid is most pronounced on the most challenging tasks: for both spatula placement and sweeping, RIGVid achieves success rates 20–25\% above the next best baseline. These results suggest that the structured 6D pose trajectory not only enables robust tracking under depth changes and occlusion but also scales to scenarios where correspondence methods fail.

\begin{figure*}[t]
    \centering

    \includegraphics[width=0.85\linewidth]{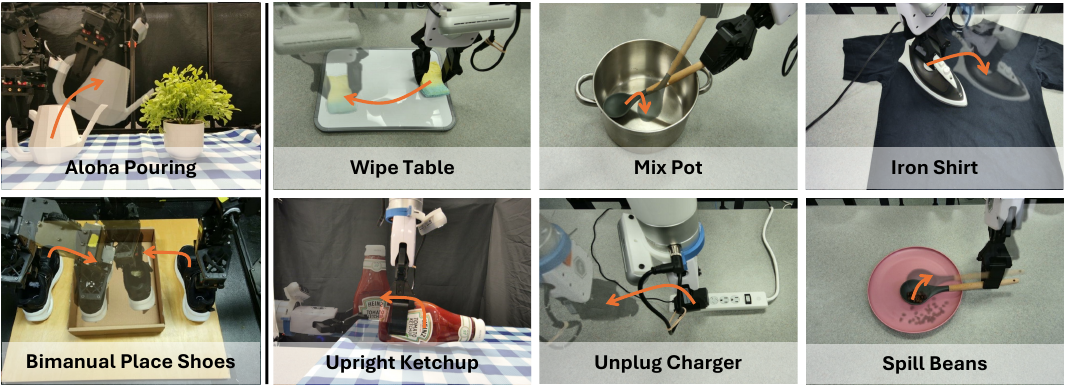}
    \captionsetup{type=figure}
    \vspace{-3pt}
    \caption{\small{\textbf{\methodshort's embodiment-agnostic capabilities and examples on solving complex, open-world tasks.} \methodshort can readily work on ALOHA setup~\cite{zhao2023learning} as shown on top left. On the bottom left, \methodshort is retargeted to the bimanual ALOHA setup. On the right, it generates trajectories for diverse manipulation tasks—including wiping, mixing, and ironing—without using any physical demonstrations.}}
    \vspace{-5pt}
    \label{fig:demos}
\end{figure*}

Visualizing the outputs in \figref{fig:placeholder} for the same generated video, we observe the intermediate predictions and resulting robot executions produced by each method. For Track2Act, the predicted tracks diverge from the true object path, leading to failed execution. Often, the track2act track prediction does not follow the true motion paths, which is the primary source of errors in our experiments. AVDC generates reasonable optical flow in individual frames, but when summed across the entire video, the resulting trajectory is often physically implausible, and the execution fails. We often found that this summing up of object flow across the video leads to small errors that accumulate over the entire video, resulting in faulty object location across the video. Gen2Act yields plausible tracks and leads to successful manipulation. We often found that tracks were accurate, and the resulting trajectory after PnP was also accurate. 4D-DPM exhibits inconsistent tracking performance. While it accurately follows the object in certain segments, the example shown reveals incorrect tracking during the first half of the episode, which ultimately causes the rollout to fail. We often found that the tracking was unstable and very jerky.  In contrast, the 6D object pose trajectories produced by RIGVid remain stable throughout the episode and closely match the actual object motion, resulting in successful execution.

\subsection{Testing Generalization}
\label{sec:robust_error_analysis}
\textbf{Embodiment-Agnostic Transfer.} 
We test \methodshort's generalizability to another embodiment by deploying it on the ALOHA robot for the pouring task (\figref{fig:demos}, top left). On this setup, it achieves 80\% success, compared to 100\% on our default xArm setup.\footnote{The slight performance drop stems primarily from camera calibration challenges, as ALOHA's arms yield less accurate pose estimates.}
\methodshort also generalizes to a bimanual setup, simultaneously placing a pair of shoes into a box using both arms (\figref{fig:demos}, bottom left).

\textbf{Extensions to Additional Tasks.} 
Besides our four main focus tasks, we also obtained promising preliminary results on a larger variety of diverse and challenging manipulation tasks shown in \figref{fig:demos} (right). These tasks include wiping, mixing, and ironing, uprighting a ketchup bottle, unplugging a charger, and rotating a spoon to spill beans. Notably, the latter three tasks involve extreme rotations, which \methodshort can handle successfully.

%% file: sec/5_conclusion.tex
\section{Conclusions}
\label{sec:conc_lim}
We introduced \method (\methodshort), the first method for robotic manipulation that works without demonstrations —- no teleoperation, no human demonstration, or expert policy rollouts. By leveraging recent advances in generative vision models and 6D pose estimation, \methodshort enables robots to execute complex tasks entirely from generated video. We extract \repr from the generated videos and retarget it to the robot, demonstrating a data-efficient approach to robotic skill acquisition. Our analysis shows a clear correlation between video quality and task success: as generation improves, \methodshort approaches real demo performance. Additionally, our comparisons with SOTA VLM-based zero-shot manipulation methods confirm that leveraging dense visual and temporal cues from generated videos yields much more reliable performance. We also show that \methodshort significantly outperforms competing trajectory extraction methods across a diverse set of visual imitation tasks, and demonstrate the robustness of our approach to environmental disturbances. Our work represents a step toward enabling robots to learn from the vast visual knowledge in generative models, reducing reliance on costly and time-consuming real-world data collection.

%% file: sec/6_acknowledgement.tex
\section{Acknowledgement}
We thank the members of the RoboPIL lab, and UIUC vision and robotics labs for their valuable discussions and feedback. Unnat would like to especially acknowledge Chen Bao, Homanga Bharadhwaj, Shikhar Bahl, and friends at CMU and Skild for their insightful conversations on learning from videos. We also thank Justin Kerr for his assistance in reproducing the 4D-DPM baseline.
This work is partially
supported by the Toyota Research Institute (TRI), the Sony
Group Corporation, Google, Dalus AI, and an Amazon Research Award, Fall 2024. This article solely
reflects the opinions and conclusions of its authors and should
not be interpreted as necessarily representing the official
policies, either expressed or implied, of the sponsors.

%% file: sec/X_suppl.tex
\clearpage
\twocolumn[
\begin{center}
  \LARGE \textbf{Appendix}
\end{center}
\vspace{1em}
]

\noindent We structure the supplement into the following subsections:
\begin{compactenum}
    \item [\ref{sec:vid_gen_prac}] Details on best practices for video generation.
    \item [\ref{sec:appendix_prompting_for_filtering}] Overview of prompt and examples of video summaries with GPT responses used for video filtering.
    \item [\ref{sec:model_free_tracking}] Results and discussion on our method's mesh-free object tracking version.
    \item [\ref{sec:remove_traj_noise}] Details on reducing noise in 6D pose rollouts for stable and realistic motion.
    \item [\ref{sec:append_baselines}] Adaptation and implementation details of baseline methods.

    \item [\ref{sec:rekep_appendix}] Comprehensive example of Rekep Predictions and Execution.

    \item [\ref{sec:gen2act_cotracker}] Discussion of limitations of Tracking using point tracks.

    \item [\ref{sec:more_robustness}] Elaboration on our method's robustness.

    \item [\ref{sec:depth_estimation_errors}] Thorough analysis of errors caused by depth estimation.

    \item [\ref{sec:mega_vs_fp}] Discussion regarding the choice between the use of MegaPose and FoundationPose, focusing on trajectory stability.

    \item [\ref{sec:appendix_compare_video_models}] Additional analysis of generated videos and human demos using VBench++ metrics.

\end{compactenum}

\section{Best Practices for Video Generation}
\label{sec:vid_gen_prac}
We found that the following practices lead to reliable video generation: (1) having a clean background without visual distractions, (2) minimizing the number of distractor objects in the scene, (3) ensuring objects are reasonably large and viewed from a natural, human-like perspective, (4) ensuring there is one clearly identifiable task that can be performed, (5) using simple and concise text prompts, and (6) setting the relevance factor to 0.7 with the negative prompt ``fast motion'' led to the most reliable video generations.

\section{Prompting for Video Filtering and Filtering Statistics}
\label{sec:appendix_prompting_for_filtering}

\begin{figure}[h]
    \centering
    \vspace{-0.5cm}
    \includegraphics[width=1\linewidth]{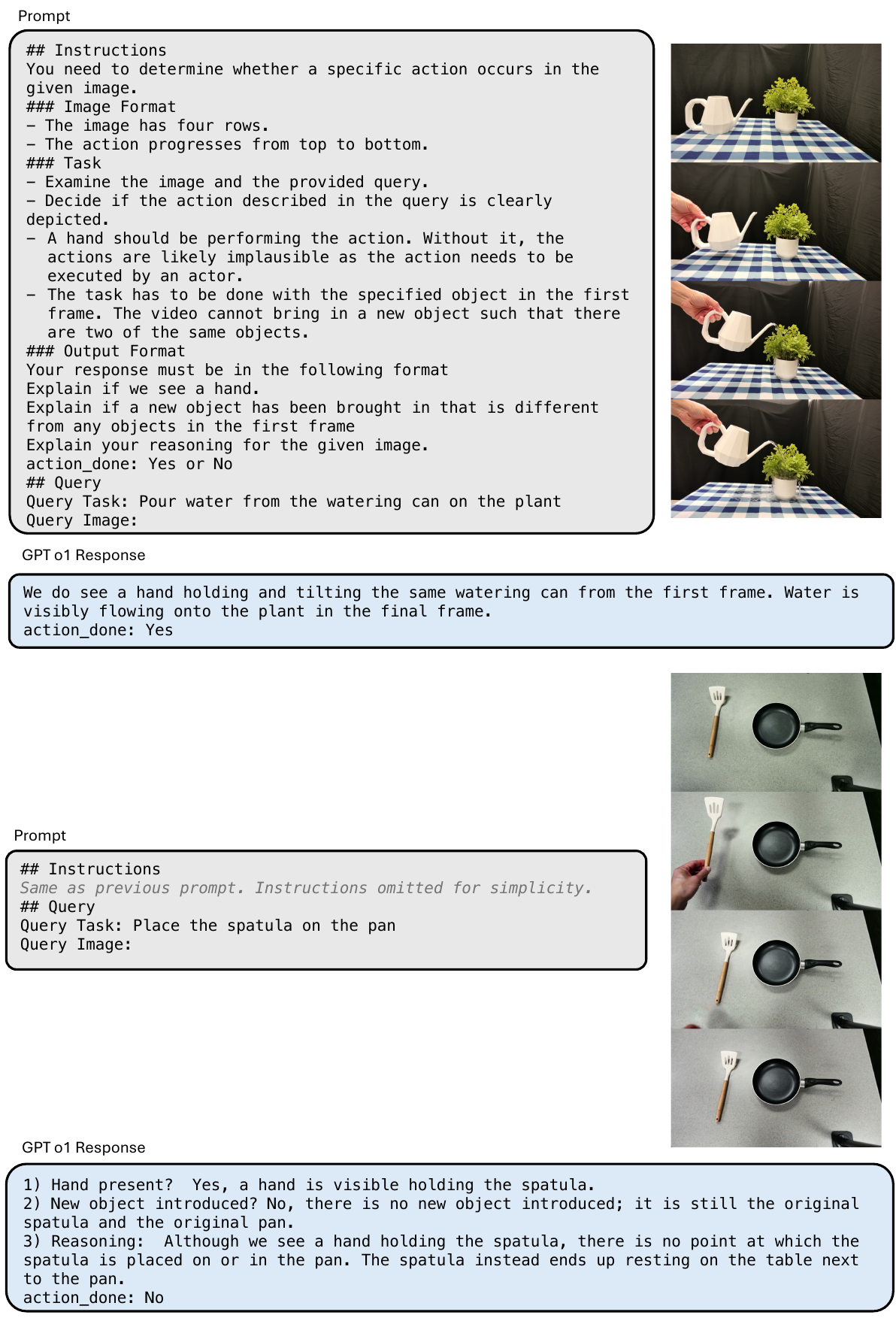}
    \captionsetup{type=figure}
    \caption{\small{\textbf{Examples of prompting GPT o1 to filter generated videos.} We sample frames from the generated video and prompt GPT o1 to assess whether the specified task is performed successfully in the video. The top example passes the filtering, while the bottom does not.}}
        \vspace{-0.5cm}

    \label{fig:prompting_example}
\end{figure}

The prompt for GPT o1-based filtering is shown in Figure \ref{fig:prompting_example}. We provide GPT o1 with the prompt, a video summary—created by vertically concatenating evenly sampled frames from the video—and the language command (e.g., "pour water"). GPT o1 then responds with "Yes" or "No" to indicate whether the task is successfully performed.

\section{Mesh-Free Object Tracking}
\label{sec:model_free_tracking}
We experiment with a mesh-free object tracking version of our method. Specifically, we use BundleSDF~\cite{wen2023bundlesdf}, which jointly performs 6-DoF object tracking and reconstruction from RGBD observations. For the \textit{pouring} task, we evaluate our method using trajectories obtained via BundleSDF over 10 trials and observe a success rate of (90\%), matching our default tracking setup. While the BundleSDF paper reports real-time capabilities, we found that its official implementation takes approximately 30 minutes to process each video in practice, which limits its applicability for real-time deployment. In contrast, our default tracker operates in real-time, enabling closed-loop execution and recovery from disturbances as discussed in \secref{sec:robust_error_analysis}. While the BundleSDF paper reports real-time capabilities, we observed significantly higher runtimes in practice with the official implementation. We expect that future advances in model-free tracking will address these efficiency bottlenecks, allowing for real-time mesh-free deployment.

\section{Smoothing Object Trajectories}
\label{sec:remove_traj_noise}
To reduce noise and jitter in the estimated object poses, we apply a moving average filter with a fixed sliding window (centered on each point) to the position and orientation components. Translations are smoothed independently along each axis, while orientation is processed similarly after converting from quaternions to rotation vectors. This approach mitigates abrupt changes, resulting in a more stable and realistic object trajectory with smoother transitions.

\section{Description of Baselines}
\label{sec:append_baselines}

\xhdr{Track2Act~\cite{bharadhwaj2024track2act}:} We adapt Track2Act's procedure to our setup preserving its core idea of object-centric trajectory estimation from point tracks. Track2Act generates a future interaction plan by predicting 2D point trajectories (using a DiT-based diffusion model) between an initial image and a goal image, then recovers a sequence of 3D object transforms via Perspective-and-Point (PnP) ~\cite{zhang2000flexible}. 

To integrate this into our pipeline, we use their published checkpoint but modify the input formulation--while the initial image remains identical to our real camera’s view, the goal image is taken from the last frame of a generated video rather than being physically captured. We then use PnP on the predicted point tracks along with the initial depth image to estimate the object's rigid motion across frames, thereby defining the end-effector trajectory. We use interpolation between consecutive poses because Track2Act generates only a sparse set of frames, and denser sampling is needed for smooth trajectory estimation and execution. However, we exclude Track2Act's closed-loop residual policy correction, focusing solely on open-loop 6D object-pose estimation and execution. This adaptation allows us to directly evaluate how well a vision-based, open-loop approach generalizes to our setting without additional corrections.

\xhdr{AVDC~\cite{ko2023learning}:} The AVDC approach models action trajectories by synthesizing a task-driven video (using a trained text-conditioned video generation model) and using optical flow from GMFlow~\cite{xu2022gmflow} to estimate dense pixel correspondences. It then reconstructs 3D object motion using an optimization step that refines pose estimates based on the tracked flow and depth information. To improve robustness, AVDC also includes a replanning mechanism that re-executes the pipeline when predicted motion stagnates.

Since the trained text-conditioned video generation model did not generalize well to our setup, we use the same generated video as in other experiments to ensure a fair comparison. While we do not employ AVDC’s replanning strategy, we predict object poses using a similar optimization framework based on flow and depth information. 

\xhdr{4D-DPM~\cite{kerr2024robot}:} 4D-DPM is designed to track 3D motion of articulated object parts from a single video. It constructs a 3D Gaussian splatting~\cite{kerbl20233d} representation of the scene to capture object features, then applies GARField~\cite{kim2024garfield} to cluster the Gaussians into discrete object components. In our adaptation, we modify this to operate on entire objects rather than individual parts. Specifically, we set the clustering parameters to treat the object as a single entity, ensuring that motion estimation is performed at the object level rather than segmenting it into multiple parts. This allows us to track and execute trajectories for the whole object.

\xhdr{Gen2Act~\cite{bharadhwaj2024gen2act}:} 
Gen2Act introduces a video-conditioned policy learning framework that first generates a human video using a video generation model from a scene image and a task description. It then extracts object tracks using BootsTAP~\cite{doersch2024bootstap}, and trains a policy using behavior cloning with an auxiliary track prediction loss and offline robot demonstrations. At inference, Gen2Act uses the generated video and the learned policy to predict robot actions.
\begin{figure}[h]
    \centering
    \includegraphics[width=1\linewidth]{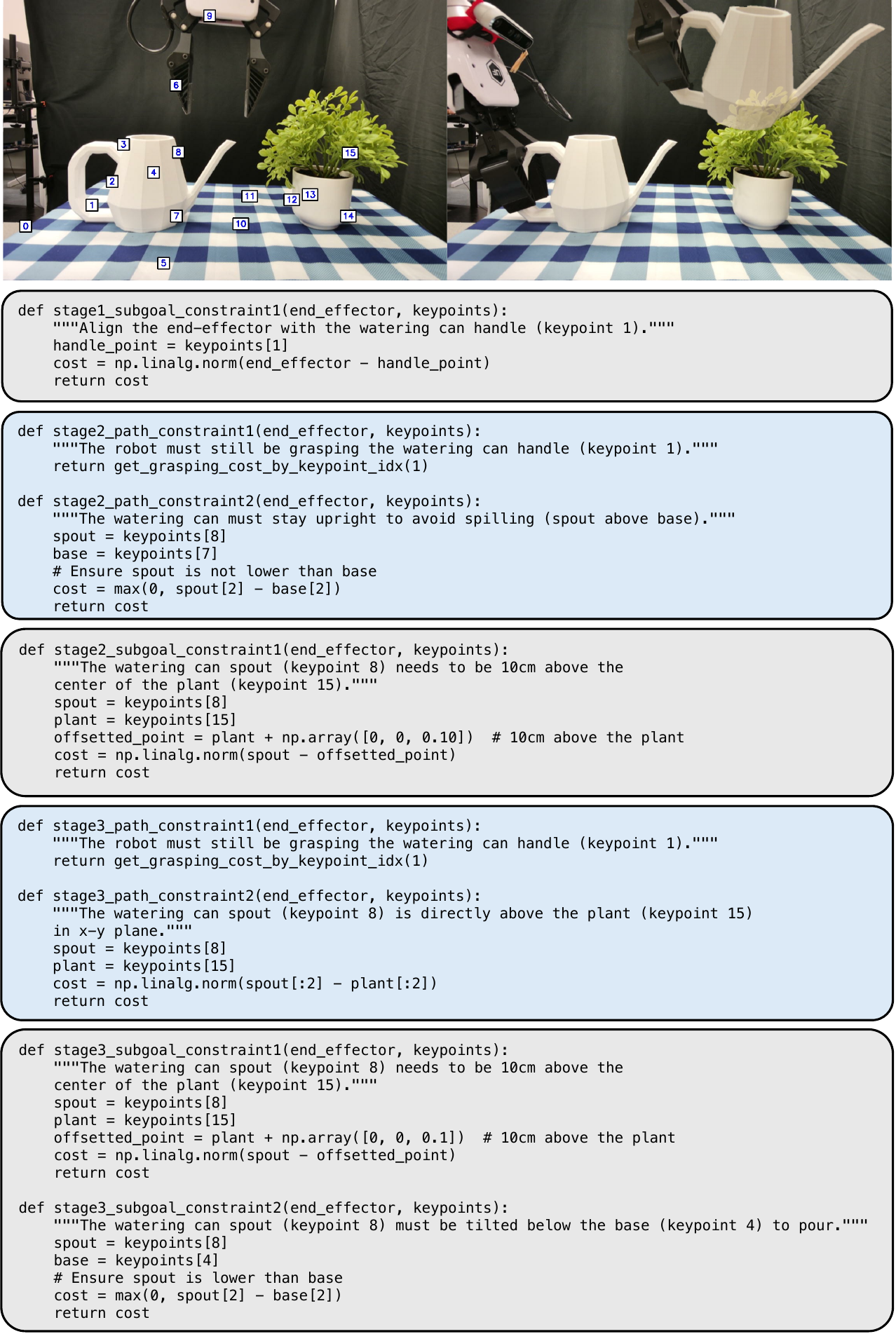}
    \captionsetup{type=figure}

    \caption{\small{\textbf{ReKep’s output for the pouring task and the resulting robot execution (top-right).} The VLM predicts to grasp at keypoint 1, move keypoint 8 above 15 and 7 during transport, and above 15 and 4 for pouring—leading to failed execution.}}
    \label{fig:rekep_example}
\end{figure}
Our approach presents a simplified adaptation of this framework that removes the need for behavior cloning, and offline demonstrations. Instead of using the extracted tracks as an auxiliary loss, we directly process them for pose estimation. To recover 3D object positions, we leverage an initial depth image corresponding to the scene image, allowing us to obtain depth values for the extracted 2D tracks. We apply RANSAC filtering to remove outlier track points and then use the Perspective-n-Point (PnP)~\cite{zhang2000flexible} to estimate the object's 6DoF pose. This adaptation preserves the core idea of leveraging video and track-based signals while eliminating the need for supervised policy learning.

\section{ReKep Predictions and Executions}
\label{sec:rekep_appendix}

\begin{figure}[t]
    \centering
    \includegraphics[width=0.7\linewidth]
    {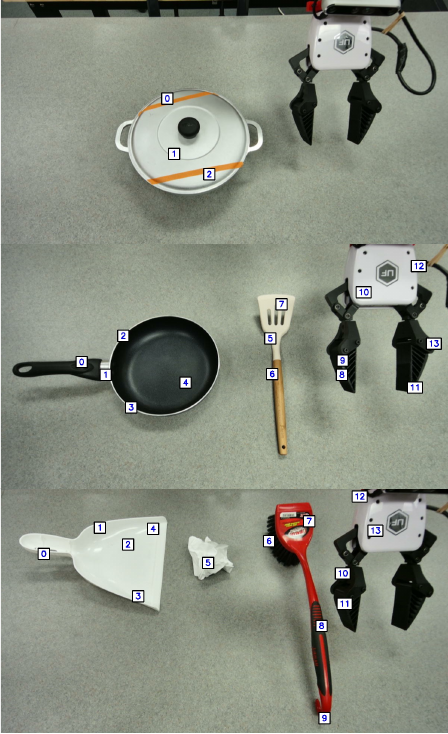}
    \captionsetup{type=figure}
    \caption{\small{\textbf{Examples of ReKep's Keypoint Locations.} The keypoint placements are often suboptimal, except for sweeping task, where the keypoints are reasonable.}}
    \label{fig:rekep_keypoints}
\end{figure}

A detailed example of ReKep's keypoint and VLM predictions for pouring task is shown in \figref{fig:rekep_example}. The VLM first predicts grasping the watering can at keypoint 1. For the transport phase, it instructs moving keypoint 8 above keypoint 15, while keeping its height above keypoint 7. For the pouring action, keypoint 8 remains above 15 (to place the spout over the plant) and above keypoint 4 (to induce tilting). The resulting robot execution fails. We attribute most ReKep failures to inaccurate keypoint predictions, as shown in \figref{fig:rekep_keypoints}. In the lid image, no keypoint appears at the lid handle. In the placing task, keypoints cluster around pan corners. For the sweeping task, the keypoints are generally well-placed, and executions succeeded. Suboptimal initial keypoints lead to inaccurate downstream VLM predictions.

\section{Limitation of Tracking with Point Tracks}
\label{sec:gen2act_cotracker}

\begin{figure}[h]
    \centering
    \includegraphics[width=\linewidth]{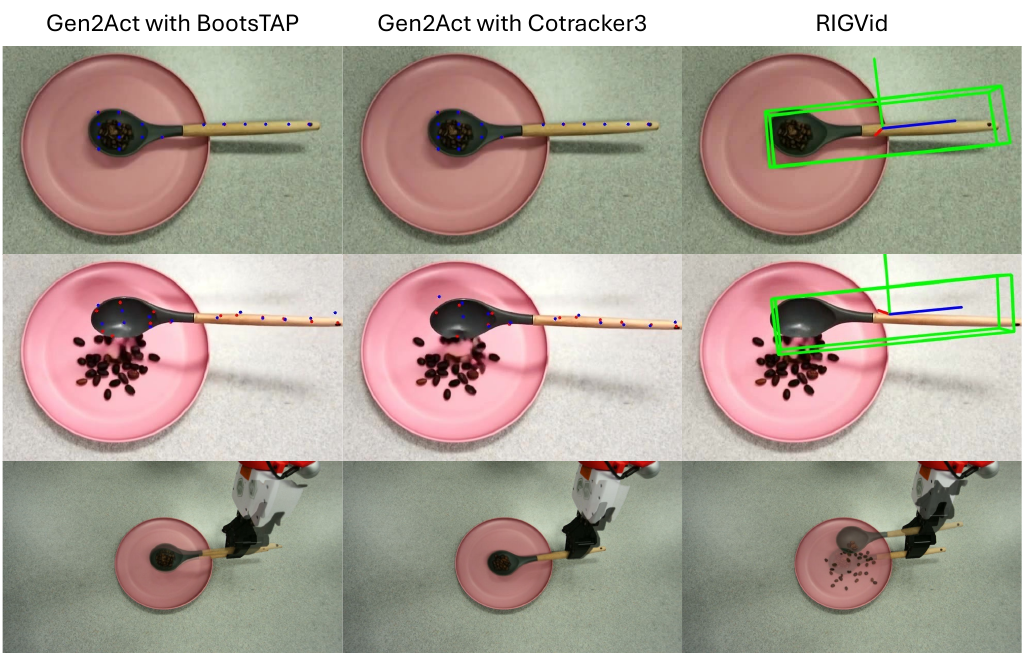}
    \captionsetup{type=figure}
    \caption{\small{\textbf{Gen2Act with BootsTAP, CoTracker, and RIGVid.} Blue points denote the tracked points used for PnP; red points represent the reprojected 3D points. For a good PnP solution, these should align, as seen in the first frame. For Gen2Act, the blue points drift significantly from the red ones in later frames, indicating failure in pose estimation due to tracking loss, which leads to failed robot execution.}}
    \vspace{-0.5cm}
    \label{fig:cotracker}
\end{figure}

All point tracks fail under extreme rotations, as initially visible points often become occluded. This is a fundamental limitation of any correspondence-based tracking method relying solely on visible surface features. We show this failure in \figref{fig:cotracker}. As the object rotates, most initial points are lost, resulting in insufficient 2D-3D correspondences to solve a stable PnP problem. This degrades pose estimation quality, leading to large drift or abrupt jumps in estimated object motion. Such instability cascades into execution errors, often causing the robot to fail the task altogether. As a result, both variants of Gen2Act—despite stronger tracking backbones like CoTracker—still fail under large out-of-plane rotations. In contrast, RIGVid's model-based 6D tracking handles these situations more robustly, as it uses full-object geometry and SE(3) filtering to maintain stable trajectories.

\section{Additional Robustness Examples}
\label{sec:more_robustness}

\begin{figure}[h]
  \centering
  \begin{minipage}{\linewidth}
    \centering
    \includegraphics[width=\linewidth]{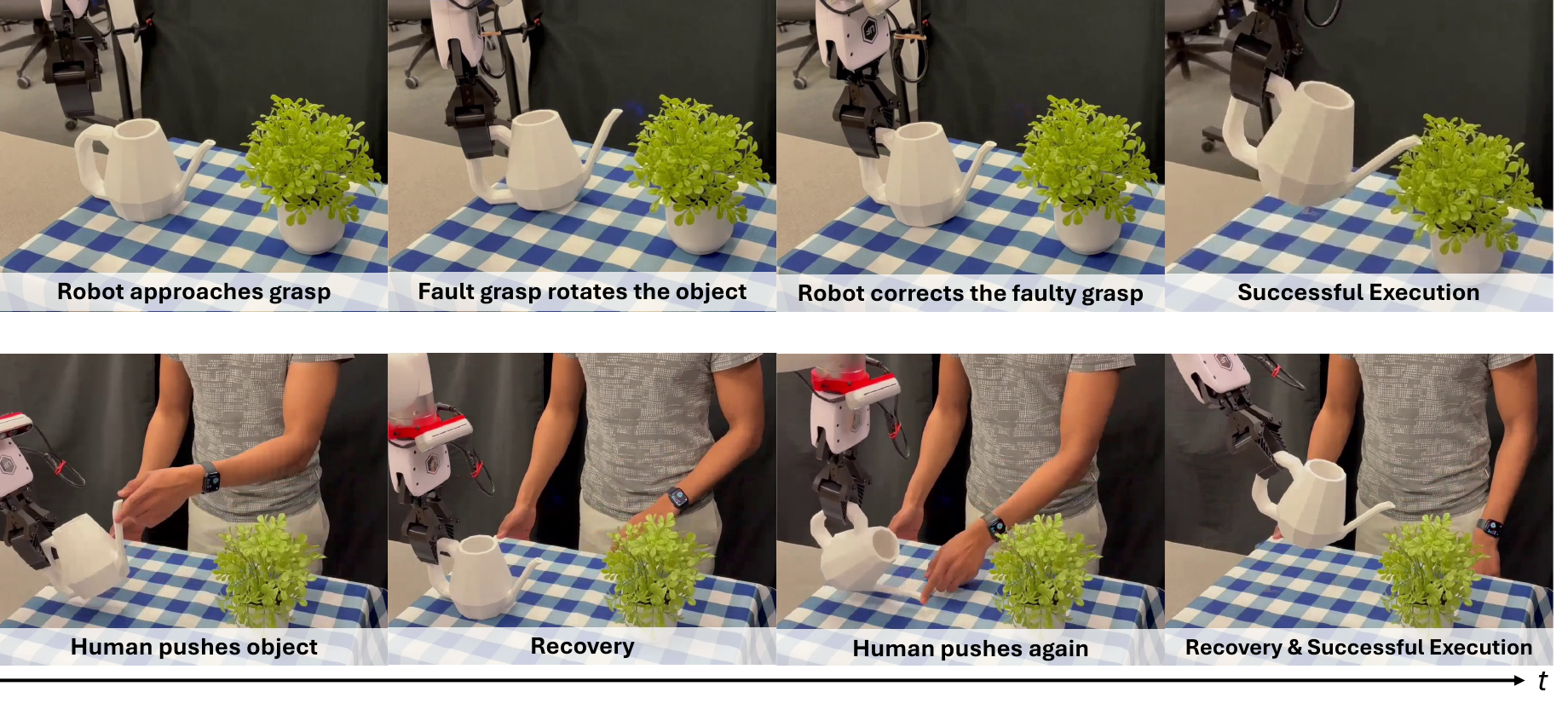}
  \end{minipage}%
  \captionsetup{type=figure}
    \caption{\small{\textbf{Additional examples of \methodshort's robustness.} In the top row, \methodshort recovers from a faulty initial grasp by reorienting the object before continuing execution. In the bottom row, it corrects for external disturbances on the object when a human pushes it mid-execution, realigning and successfully completing the task.}}
  \label{fig:more_robustness}
\end{figure}

Examples of \methodshort's robustness are shown in \figref{fig:more_robustness}. In the first row, the robot grasps the object, but due to a misaligned grasp, the object rotates unexpectedly. The robot compensates by rotating it back to the correct orientation and then resumes the planned trajectory, completing the task successfully. In the bottom row, a human perturbs the object during execution while it is held by the robot. \methodshort detects the resulting change in the relative transformation and automatically re-aligns the object before continuing. When the human intervenes a second time, \methodshort again corrects the deviation, resulting in successful task completion.

\section{Errors from Depth Estimation}
\label{sec:depth_estimation_errors}

\begin{figure}[h]
  \centering
  \begin{minipage}{\linewidth}
    \centering
    \includegraphics[width=\linewidth]{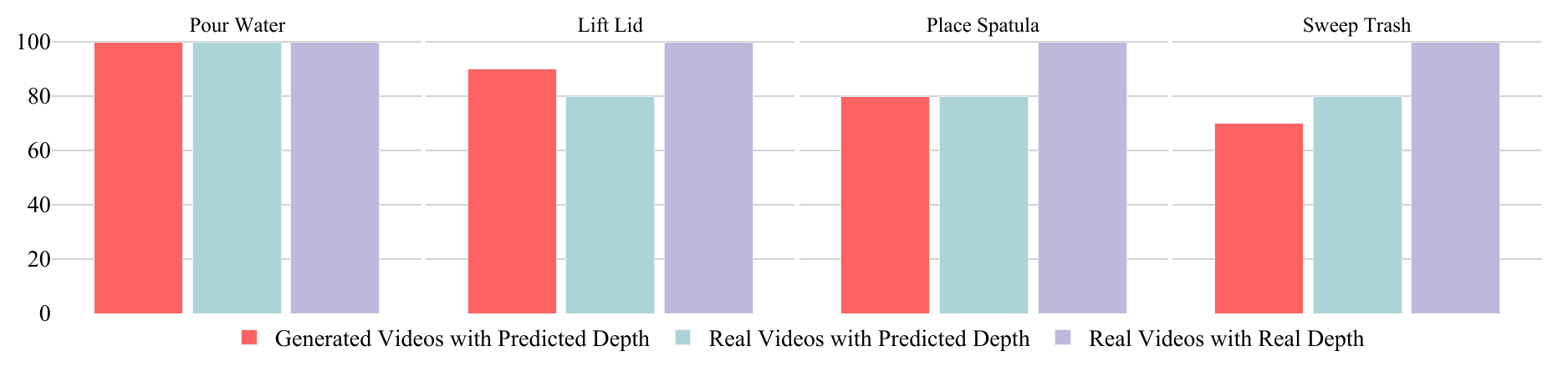}
  \end{minipage}%
  \captionsetup{type=figure}
  \caption{\small{\textbf{ Impact of Depth Estimation Errors on \algabrvname performance.} Errors in monocular depth estimation result in worse performance of generated and real videos. \methodshort achieves perfect success across all tasks with real videos and real depth.}}
  \label{fig:depth_pred_bar}
  \vspace{-0.2cm}
\end{figure}

\begin{figure}[t]
    \centering
    \begin{subfigure}{0.8\linewidth}
        \centering
        \small{(a) Generated Video} \\
        \includegraphics[width=\linewidth]{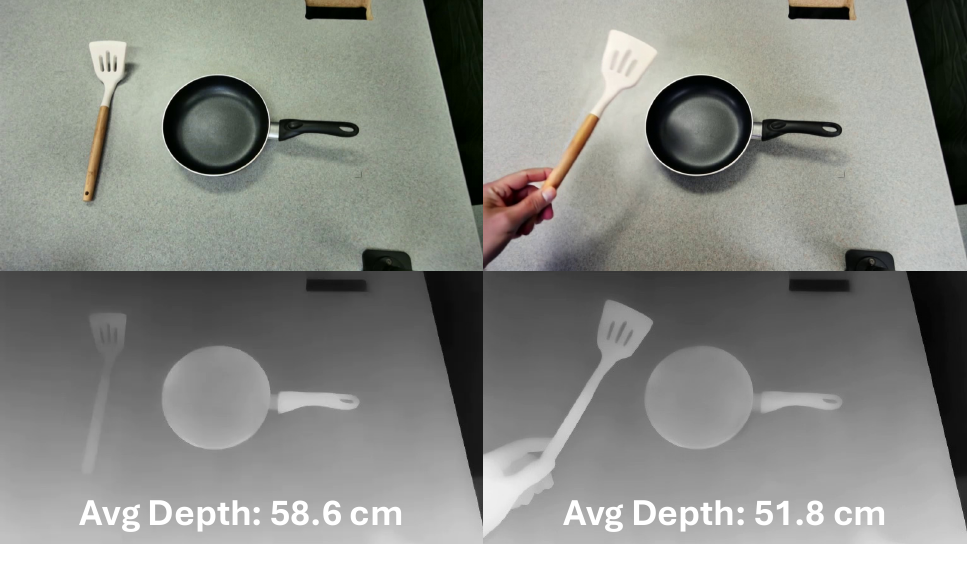}
    \end{subfigure}

    \vspace{0.3cm} %

    \begin{subfigure}{0.8\linewidth}
        \centering
        \small{(b) Real Video} \\
        \includegraphics[width=\linewidth]{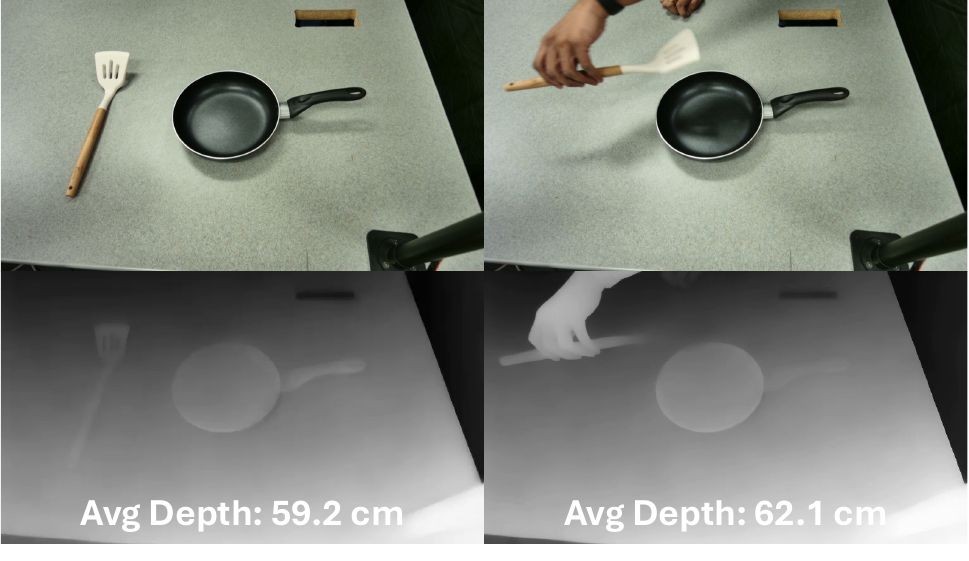}
    \end{subfigure}

    \caption{\small{\textbf{Errors in Monocular Depth Estimation.} In the generated video (top), the depth of the spatula changes only slightly despite a large visual change. In the real video (bottom), the spatula's head is predicted to lie farther away, contradicting the visual appearance.}}
    \label{fig:depth_err}
    \vspace{-0.5cm}
\end{figure}

\begin{figure}[t]
    \centering
    \begin{subfigure}{0.8\linewidth}
        \centering
        \small{(a) Generated Video} \\
        \includegraphics[width=\linewidth]{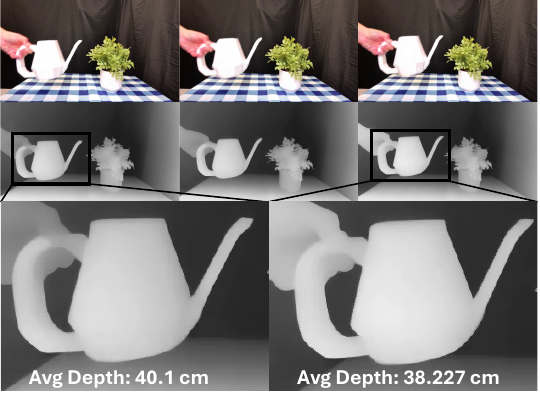}
    \end{subfigure}

    \vspace{0.3cm} %

    \begin{subfigure}{0.8\linewidth}
        \centering
        \small{(b) Real Video} \\
        \includegraphics[width=\linewidth]{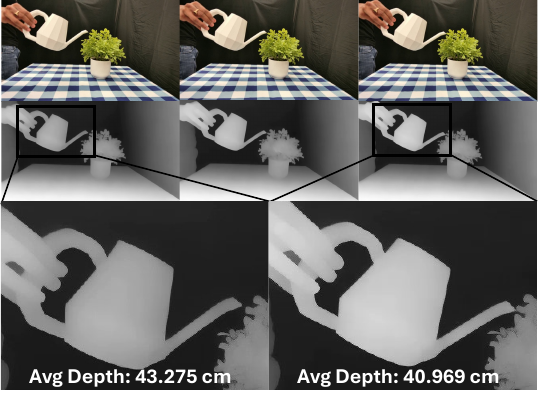}
    \end{subfigure}

    \caption{\small{\textbf{Flickering in Depth Prediction.} We show three consecutive frames of the video and its corresponding predicted depth. The depth of the watering can change noticeably across frames—appearing significantly whiter in the third frame despite minimal actual motion. We observe this behavior in both generated and real videos.}}
    \label{fig:depth_flickering}
\end{figure}

In \figref{fig:depth_pred_bar}, we isolate the impact of depth estimation errors. Robot executions on real videos with real depth (captured using an RGBD camera) achieve a 100\% success rate, whereas executions from real videos with generated depth result in an 85\% average success. Similarly, executions from Kling V1.6-generated videos with generated depth also achieve 85\% success, suggesting that the primary source of error lies in monocular depth estimation. Upon inspection, we observe two common undesirable behaviors in the predicted depth: inaccurate depth values and temporal flickering. An example of inaccurate depth is shown in \figref{fig:depth_err}. In the generated video, when the spatula is brought close to the camera, the depth changes by only 6.8\,cm, which is visibly inconsistent with the video and likely much smaller than the real-world change. Inaccuracies also occur in real videos, as shown in the figure—the head of the spatula is estimated to be far from the camera, despite appearing close, revealing another failure mode in monocular depth estimation. Flickering is shown in \figref{fig:depth_flickering}. Although the position of the watering can relative to the camera remains nearly unchanged across three consecutive frames, the estimated depth varies significantly. The zoomed-in region on the right shows the can appearing much whiter than on the left, indicating a substantial change in predicted depth. The average depth of the can changes from 40.1\,cm to 38.2\,cm--a 1.9\,cm difference over just 0.066 seconds--which is physically implausible for the generated video. We find similar flickering behavior in real videos as well, where the depth changes from 43.2\,cm to 40.9\,cm in the given example--a 2.3\,cm difference. Since errors in the generated depth are the main source of failure, we also tested removing it entirely by estimating object pose directly from the RGB frames of the generated video using MegaPose. However, this approach leads to even more unstable and noisy trajectories, as detailed in \appref{sec:mega_vs_fp}.

\section{Choice between MegaPose and FoundationPose}
\label{sec:mega_vs_fp}

We compare trajectory stability from MegaPose~\cite{labbe2022megapose} and FoundationPose~\cite{wen2023foundationpose} by computing the translational and rotational RMS jitter. For each method, we apply a Gaussian smoothing filter ($\sigma=2$ frames) to the raw SE(3) pose sequences, compute the residual between original and smoothed trajectories, and then calculate:
\[
\text{jitter}_{\text{trans}} = \sqrt{\frac{1}{N}\sum_{t=1}^N \|\Delta \mathbf{t}_t\|^2}, \quad
\text{jitter}_{\text{rot}} = \sqrt{\frac{1}{N}\sum_{t=1}^N \theta_t^2},
\]
where \(\Delta \mathbf{t}_t\) is the translational residual at frame \(t\), and \(\theta_t\) is the angular magnitude (in radians) of the relative rotation \(R_{\text{smooth}}^{-1}R_{\text{raw}}\), converted to degrees. Metrics are averaged over ten pouring trajectories from generated videos.

MegaPose yields an average translational RMS jitter of 0.0045m and rotational RMS jitter of 37.47°, whereas FoundationPose achieves 0.0029m translational and 14.31° rotational jitter. This demonstrate that FoundationPose produces significantly smoother and more stable trajectories. Additionally, it allows for real-time tracking during the execution, making \methodshort robust to external disturbances.

\section{Comparing Video Generative Models}
\label{sec:appendix_compare_video_models}

\begin{figure*}[h]
    \centering
    \includegraphics[width=0.82\linewidth]{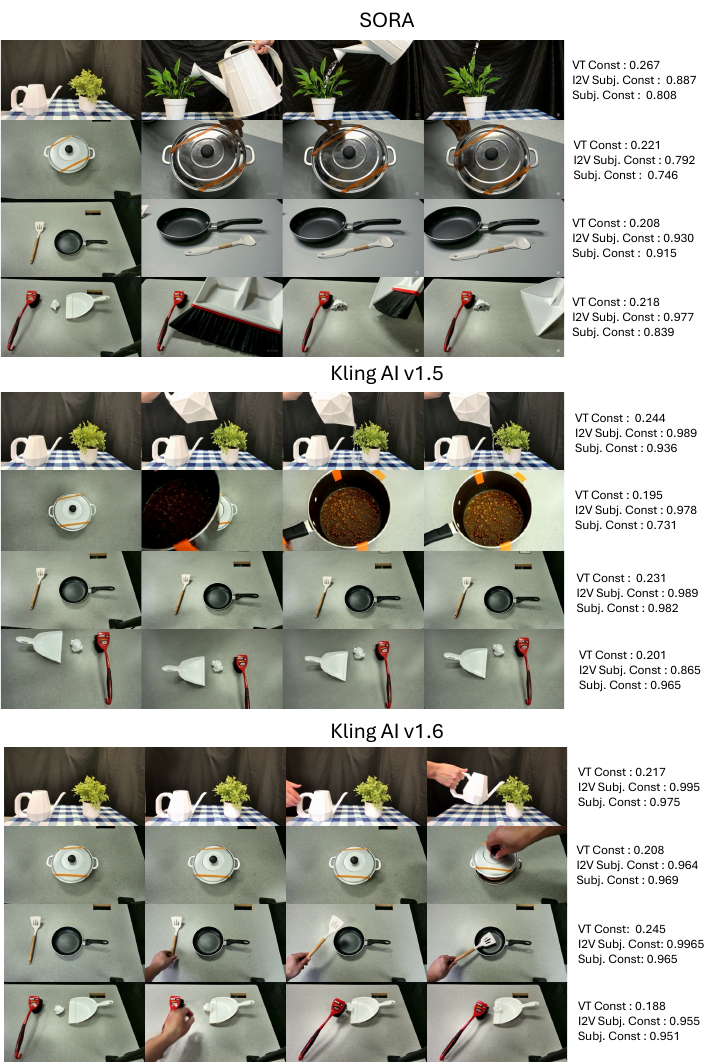}
    \captionsetup{type=figure}
    \caption{\small{\textbf{Qualitative Comparison of Different Video Generative Models.} Videos from the three video generation models are shown using evenly sampled frames, along with VBench++~\cite{huang2024vbench++} metrics: video-text consistency, image-to-video subject consistency, and subject consistency. Kling v1.6 scores highest on these metrics, followed by Kling v1.5 and then Sora.}} 
    \label{fig:qualitative_model_compare}
\end{figure*}

To further assess video quality, we report VBench++ \cite{huang2024vbench++} metrics in Table \ref{tab:video_metrics_different_models} and explain them below. The numbers in the table are scaled 100$\times$ for easier interpretation. We collect these metrics on 40 randomly selected and unfiltered videos per model, 10 for each of the four tasks. Kling v1.6 outperformed the other models on most metrics but performed similarly or worse in video-text consistency and dynamic degree. Human evaluations discussed in \secref{sec:video_model_comparison} suggest that the video-text consistency and I2V subject consistency are not reliable indicators of whether a generated video correctly follows a given command. Sora scored high on dynamic degree, likely due to its tendency to drastically alter the scene, resulting in exceptionally large motions. Generated videos from these models and their corresponding metrics are shown in \figref{fig:qualitative_model_compare} and further details on these metrics can be found the next section.

\vspace{0.5em} \noindent\textbf{VBench++ Metric Definitions:}

\noindent$\bullet$~\textbf{Subject Consistency.} Subject consistency describes whether subjects' appearance remain consistent, which is computed by DINOv1 \cite{caron2021emerging} similarities across video frames.

\noindent$\bullet$~\textbf{Background Consistency.} Background temporal consistency by CLIP \cite{radford2021learningtransferablevisualmodels} similarities across frames. 

\noindent$\bullet$~\textbf{Motion Smoothness.} Evaluates smoothness of videos by utilizing video frame interpolation model AMT \cite{licvpr23amt}.

\noindent$\bullet$~\textbf{Dynamic Degree.} Describes whether the video contains large motions as a binary metric.

\noindent$\bullet$~\textbf{Aesthetic Quality.} Human perceived artistic and beauty value such as photo-realism, layout and color harmony.

\noindent$\bullet$~\textbf{Imaging Quality.} Assesses the presence of distortion in a video, such as noisiness, blurriness, and over-exposure.

\noindent$\bullet$~\textbf{Video-Text Consistency.} Text-to-video alignment score calculated by ViCLIP \cite{wang2023internvid}.

\noindent$\bullet$~\textbf{I2V Subject Consistency.} Similarity between subjects in input image and each video frame, as well as similarity between consecutive frames. Features are extracted from DINOv1 \cite{caron2021emerging}.

\begin{table}[h]
    \centering
    \resizebox{0.5\textwidth}{!}{
    \begin{tabular}{lcccc}
        \toprule
        \multirow{2}{*}{\textbf{Metrics}} & \multicolumn{3}{c}{\textbf{Video Generation Models}} & \multirow{2}{*}{\begin{tabular}[c]{@{}c@{}}\textbf{Human} \\ \textbf{Demos}\end{tabular}}\\
        \cmidrule(lr){2-4}
        & Kling V1.6 & Kling V1.5 & Sora &  \\
        \midrule
        Subject Consistency  & \textbf{96.34} & 91.66 & 83.09 & 94.91 \\
        Background Consistency & \textbf{96.64} & 93.97 & 89.34 & 95.00 \\
        Motion Smoothness  & \textbf{99.68} & 99.57 & 99.06 & 99.51\\
        Dynamic Degree  & 52.5 & 57.5 & \textbf{70.0} & 80.0 \\
        Aesthetic Quality  & \textbf{51.75} & 49.77 & 46.22 & 49.30 \\
        Imaging Quality  & \textbf{72.80} & 71.48 & 68.68 & 72.52 \\
        Video-Text Consistency & 22.01 & \textbf{22.61} & 21.42  & 21.57\\
        I2V Subject Consistency  & \textbf{97.88} & 95.96 & 89.09 & 97.89\\
        \bottomrule
    \end{tabular}
    }
    \vspace{0.3cm}
    \caption{\small{\textbf{Video generation quality metrics for real human demonstration videos and different models.} Higher values indicate better quality. Kling v1.6 performs comparably to or surpasses other models on most metrics.}}
    \label{tab:video_metrics_different_models}
\end{table}

\newpage

\newpage
\begin{figure*}[h]
    \centering
     \includegraphics[width=0.7\linewidth]{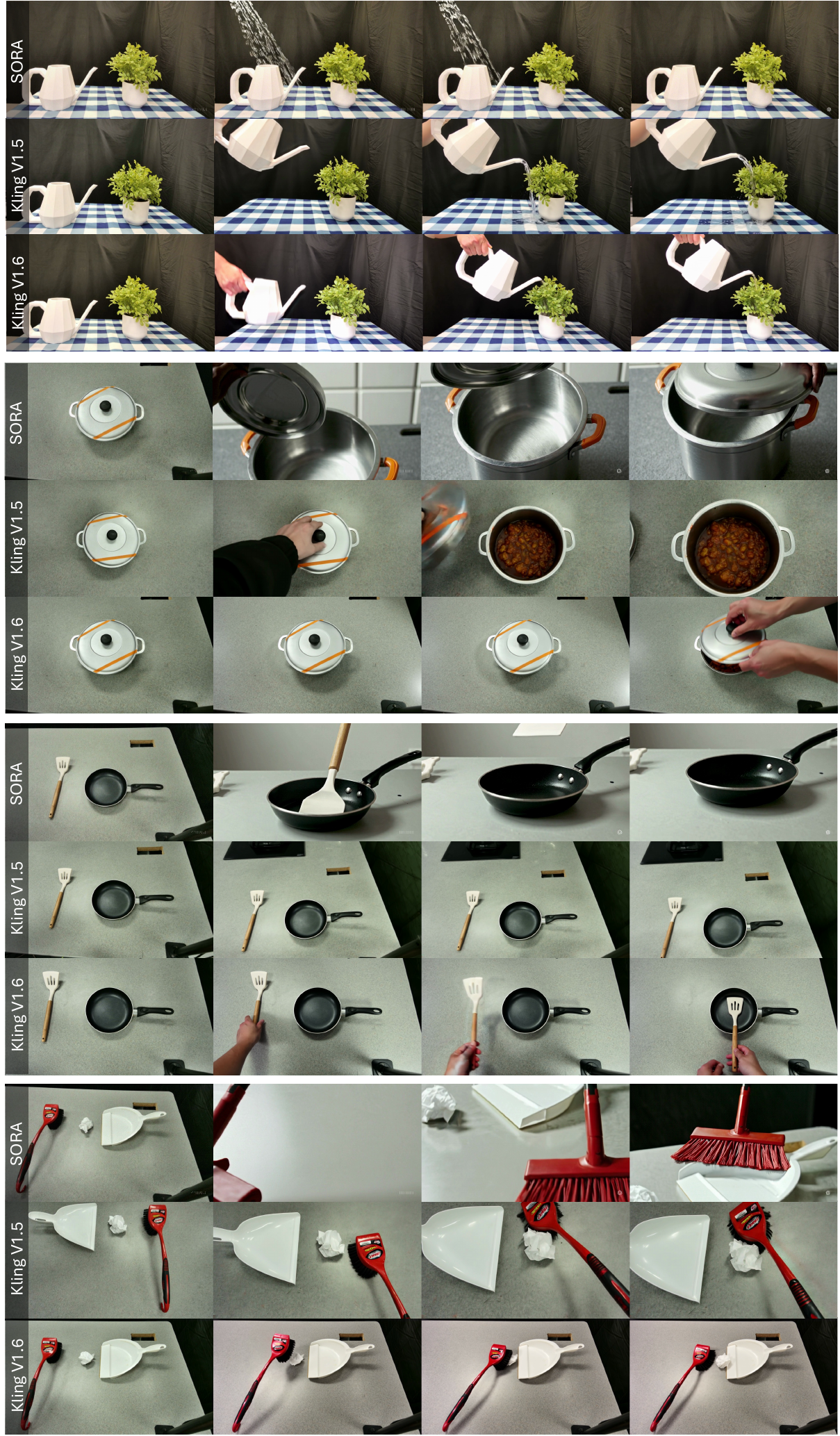}
    \captionsetup{type=figure}
    \caption{\small{\textbf{Qualitative comparison of video generation.} Sora-generated videos often alter the scene layout and objects. Kling V1.5 produces more plausible results but includes physically implausible elements. Kling V1.6 better preserves scene fidelity and closely follows the human command.}}

    \label{fig:large_qualitative_model_compare}
\end{figure*}

%% file: main.bbl
\begin{thebibliography}{138}
\providecommand{\natexlab}[1]{#1}
\providecommand{\url}[1]{\texttt{#1}}
\expandafter\ifx\csname urlstyle\endcsname\relax
  \providecommand{\doi}[1]{doi: #1}\else
  \providecommand{\doi}{doi: \begingroup \urlstyle{rm}\Url}\fi

\bibitem[Kli(2024)]{KlingAI2024}
Kling ai.
\newblock \url{https://www.klingai.com/}, 2024.
\newblock Accessed: 2024-02-10.

\bibitem[Achiam et~al.(2023)Achiam, Adler, Agarwal, Ahmad, Akkaya, Aleman, Almeida, Altenschmidt, Altman, Anadkat, et~al.]{achiam2023gpt}
Josh Achiam, Steven Adler, Sandhini Agarwal, Lama Ahmad, Ilge Akkaya, Florencia~Leoni Aleman, Diogo Almeida, Janko Altenschmidt, Sam Altman, Shyamal Anadkat, et~al.
\newblock Gpt-4 technical report.
\newblock \emph{arXiv preprint arXiv:2303.08774}, 2023.

\bibitem[Ajay et~al.(2023)Ajay, Han, Du, Li, Gupta, Jaakkola, Tenenbaum, Kaelbling, Srivastava, and Agrawal]{ajay2023compositional}
Anurag Ajay, Seungwook Han, Yilun Du, Shuang Li, Abhi Gupta, Tommi Jaakkola, Josh Tenenbaum, Leslie Kaelbling, Akash Srivastava, and Pulkit Agrawal.
\newblock Compositional foundation models for hierarchical planning.
\newblock \emph{Advances in Neural Information Processing Systems}, 36:\penalty0 22304--22325, 2023.

\bibitem[Albaba et~al.(2025)Albaba, Li, Diomataris, Taheri, Krause, and Black]{albaba2025nil}
Mert Albaba, Chenhao Li, Markos Diomataris, Omid Taheri, Andreas Krause, and Michael Black.
\newblock Nil: No-data imitation learning by leveraging pre-trained video diffusion models.
\newblock \emph{arXiv preprint arXiv:2503.10626}, 2025.

\bibitem[Argus et~al.(2020)Argus, Hermann, Long, and Brox]{argus2020flowcontrol}
Max Argus, Lukas Hermann, Jon Long, and Thomas Brox.
\newblock Flowcontrol: Optical flow based visual servoing.
\newblock In \emph{2020 IEEE/RSJ International Conference on Intelligent Robots and Systems (IROS)}, pages 7534--7541. IEEE, 2020.

\bibitem[Ausserlechner et~al.(2024)Ausserlechner, Haberger, Thalhammer, Weibel, and Vincze]{ausserlechner2024zs6d}
Philipp Ausserlechner, David Haberger, Stefan Thalhammer, Jean-Baptiste Weibel, and Markus Vincze.
\newblock Zs6d: Zero-shot 6d object pose estimation using vision transformers.
\newblock In \emph{2024 IEEE International Conference on Robotics and Automation (ICRA)}, pages 463--469. IEEE, 2024.

\bibitem[Bahety et~al.(2024)Bahety, Mandikal, Abbatematteo, and Mart{\'\i}n-Mart{\'\i}n]{bahety2024screwmimic}
Arpit Bahety, Priyanka Mandikal, Ben Abbatematteo, and Roberto Mart{\'\i}n-Mart{\'\i}n.
\newblock Screwmimic: Bimanual imitation from human videos with screw space projection.
\newblock \emph{arXiv preprint arXiv:2405.03666}, 2024.

\bibitem[Bahl et~al.(2022)Bahl, Gupta, and Pathak]{bahl2022human}
Shikhar Bahl, Abhinav Gupta, and Deepak Pathak.
\newblock Human-to-robot imitation in the wild.
\newblock \emph{arXiv preprint arXiv:2207.09450}, 2022.

\bibitem[Bahl et~al.(2023)Bahl, Mendonca, Chen, Jain, and Pathak]{bahl2023affordances}
Shikhar Bahl, Russell Mendonca, Lili Chen, Unnat Jain, and Deepak Pathak.
\newblock Affordances from human videos as a versatile representation for robotics.
\newblock In \emph{Proceedings of the IEEE/CVF Conference on Computer Vision and Pattern Recognition}, pages 13778--13790, 2023.

\bibitem[Baker et~al.(2022)Baker, Akkaya, Zhokov, Huizinga, Tang, Ecoffet, Houghton, Sampedro, and Clune]{baker2022video}
Bowen Baker, Ilge Akkaya, Peter Zhokov, Joost Huizinga, Jie Tang, Adrien Ecoffet, Brandon Houghton, Raul Sampedro, and Jeff Clune.
\newblock Video pretraining (vpt): Learning to act by watching unlabeled online videos.
\newblock \emph{Advances in Neural Information Processing Systems}, 35:\penalty0 24639--24654, 2022.

\bibitem[Bansal et~al.(2024)Bansal, Lin, Xie, Zong, Yarom, Bitton, Jiang, Sun, Chang, and Grover]{bansal2024videophy}
Hritik Bansal, Zongyu Lin, Tianyi Xie, Zeshun Zong, Michal Yarom, Yonatan Bitton, Chenfanfu Jiang, Yizhou Sun, Kai-Wei Chang, and Aditya Grover.
\newblock Videophy: Evaluating physical commonsense for video generation.
\newblock \emph{arXiv preprint arXiv:2406.03520}, 2024.

\bibitem[Barcellona et~al.(2024)Barcellona, Zadaianchuk, Allegro, Papa, Ghidoni, and Gavves]{barcellona2024dream}
Leonardo Barcellona, Andrii Zadaianchuk, Davide Allegro, Samuele Papa, Stefano Ghidoni, and Efstratios Gavves.
\newblock Dream to manipulate: Compositional world models empowering robot imitation learning with imagination.
\newblock \emph{arXiv preprint arXiv:2412.14957}, 2024.

\bibitem[Bharadhwaj et~al.(2023)Bharadhwaj, Gupta, Tulsiani, and Kumar]{bharadhwaj2023zero}
Homanga Bharadhwaj, Abhinav Gupta, Shubham Tulsiani, and Vikash Kumar.
\newblock Zero-shot robot manipulation from passive human videos.
\newblock \emph{arXiv preprint arXiv:2302.02011}, 2023.

\bibitem[Bharadhwaj et~al.(2024{\natexlab{a}})Bharadhwaj, Dwibedi, Gupta, Tulsiani, Doersch, Xiao, Shah, Xia, Sadigh, and Kirmani]{bharadhwaj2024gen2act}
Homanga Bharadhwaj, Debidatta Dwibedi, Abhinav Gupta, Shubham Tulsiani, Carl Doersch, Ted Xiao, Dhruv Shah, Fei Xia, Dorsa Sadigh, and Sean Kirmani.
\newblock Gen2act: Human video generation in novel scenarios enables generalizable robot manipulation.
\newblock \emph{arXiv preprint arXiv:2409.16283}, 2024{\natexlab{a}}.

\bibitem[Bharadhwaj et~al.(2024{\natexlab{b}})Bharadhwaj, Mottaghi, Gupta, and Tulsiani]{bharadhwaj2024track2act}
Homanga Bharadhwaj, Roozbeh Mottaghi, Abhinav Gupta, and Shubham Tulsiani.
\newblock Track2act: Predicting point tracks from internet videos enables diverse zero-shot robot manipulation, 2024{\natexlab{b}}.

\bibitem[Brooks et~al.(2024)Brooks, Peebles, Holmes, DePue, Guo, Jing, Schnurr, Taylor, Luhman, Luhman, Ng, Wang, and Ramesh]{videoworldsimulators2024}
Tim Brooks, Bill Peebles, Connor Holmes, Will DePue, Yufei Guo, Li Jing, David Schnurr, Joe Taylor, Troy Luhman, Eric Luhman, Clarence Ng, Ricky Wang, and Aditya Ramesh.
\newblock Video generation models as world simulators.
\newblock 2024.

\bibitem[Cai and Reid(2020)]{cai2020reconstruct}
Ming Cai and Ian Reid.
\newblock Reconstruct locally, localize globally: A model free method for object pose estimation.
\newblock In \emph{Proceedings of the IEEE/CVF Conference on Computer Vision and Pattern Recognition}, pages 3153--3163, 2020.

\bibitem[Calinon(2016)]{calinon2016tutorial}
Sylvain Calinon.
\newblock A tutorial on task-parameterized movement learning and retrieval.
\newblock \emph{Intelligent service robotics}, 9:\penalty0 1--29, 2016.

\bibitem[Caraffa et~al.(2024)Caraffa, Boscaini, Hamza, and Poiesi]{caraffa2024freeze}
Andrea Caraffa, Davide Boscaini, Amir Hamza, and Fabio Poiesi.
\newblock Freeze: Training-free zero-shot 6d pose estimation with geometric and vision foundation models.
\newblock \emph{European Conference on Computer Vision (ECCV)}, 2024.

\bibitem[Caron et~al.(2021)Caron, Touvron, Misra, J\'egou, Mairal, Bojanowski, and Joulin]{caron2021emerging}
Mathilde Caron, Hugo Touvron, Ishan Misra, Herv\'e J\'egou, Julien Mairal, Piotr Bojanowski, and Armand Joulin.
\newblock Emerging properties in self-supervised vision transformers.
\newblock In \emph{Proceedings of the International Conference on Computer Vision (ICCV)}, 2021.

\bibitem[Chane-Sane et~al.(2023)Chane-Sane, Schmid, and Laptev]{chane2023learning}
Elliot Chane-Sane, Cordelia Schmid, and Ivan Laptev.
\newblock Learning video-conditioned policies for unseen manipulation tasks.
\newblock In \emph{2023 IEEE International Conference on Robotics and Automation (ICRA)}, pages 909--916. IEEE, 2023.

\bibitem[Chang et~al.(2020)Chang, Gupta, and Gupta]{chang2020semantic}
Matthew Chang, Arjun Gupta, and Saurabh Gupta.
\newblock Semantic visual navigation by watching youtube videos.
\newblock In \emph{NeurIPS}, 2020.

\bibitem[Chang et~al.(2023)Chang, Gervet, Khanna, Yenamandra, Shah, Min, Shah, Paxton, Gupta, Batra, Mottaghi, Malik, and Chaplot]{chang2023goatthing}
Matthew Chang, Theophile Gervet, Mukul Khanna, Sriram Yenamandra, Dhruv Shah, So~Yeon Min, Kavit Shah, Chris Paxton, Saurabh Gupta, Dhruv Batra, Roozbeh Mottaghi, Jitendra Malik, and Devendra~Singh Chaplot.
\newblock Goat: Go to any thing.
\newblock \emph{arXiv preprint arXiv:2311.06430}, 2023.

\bibitem[Cheng et~al.(2024)Cheng, Ji, Chen, Yang, Yang, and Wang]{cheng2024expressive}
Xuxin Cheng, Yandong Ji, Junming Chen, Ruihan Yang, Ge Yang, and Xiaolong Wang.
\newblock Expressive whole-body control for humanoid robots.
\newblock \emph{arXiv preprint arXiv:2402.16796}, 2024.

\bibitem[Choi et~al.(2020)Choi, Pan, and Kim]{choi2020nonparametric}
Sungjoon Choi, Matthew~KXJ Pan, and Joohyung Kim.
\newblock Nonparametric motion retargeting for humanoid robots on shared latent space.
\newblock In \emph{Robotics: science and systems}, 2020.

\bibitem[Dasari and Gupta(2021)]{dasari2021transformers}
Sudeep Dasari and Abhinav Gupta.
\newblock Transformers for one-shot visual imitation.
\newblock In \emph{Conference on Robot Learning}, pages 2071--2084. PMLR, 2021.

\bibitem[Dasari et~al.(2023)Dasari, Srirama, Jain, and Gupta]{dasari2023unbiased}
Sudeep Dasari, Mohan~Kumar Srirama, Unnat Jain, and Abhinav Gupta.
\newblock An unbiased look at datasets for visuo-motor pre-training.
\newblock In \emph{Conference on Robot Learning}, pages 1183--1198. PMLR, 2023.

\bibitem[Doersch et~al.(2024)Doersch, Yang, Gokay, Luc, Koppula, Gupta, Heyward, Goroshin, Carreira, and Zisserman]{doersch2024bootstap}
Carl Doersch, Yi Yang, Dilara Gokay, Pauline Luc, Skanda Koppula, Ankush Gupta, Joseph Heyward, Ross Goroshin, Jo{\~a}o Carreira, and Andrew Zisserman.
\newblock Bootstap: Bootstrapped training for tracking-any-point.
\newblock \emph{arXiv preprint arXiv:2402.00847}, 2024.

\bibitem[Du et~al.(2023)Du, Yang, Florence, Xia, Wahid, Ichter, Sermanet, Yu, Abbeel, Tenenbaum, et~al.]{du2023video}
Yilun Du, Mengjiao Yang, Pete Florence, Fei Xia, Ayzaan Wahid, Brian Ichter, Pierre Sermanet, Tianhe Yu, Pieter Abbeel, Joshua~B Tenenbaum, et~al.
\newblock Video language planning.
\newblock \emph{arXiv preprint arXiv:2310.10625}, 2023.

\bibitem[Du et~al.(2024)Du, Yang, Dai, Dai, Nachum, Tenenbaum, Schuurmans, and Abbeel]{du2024learning}
Yilun Du, Sherry Yang, Bo Dai, Hanjun Dai, Ofir Nachum, Josh Tenenbaum, Dale Schuurmans, and Pieter Abbeel.
\newblock Learning universal policies via text-guided video generation.
\newblock \emph{Advances in Neural Information Processing Systems}, 36, 2024.

\bibitem[Fang et~al.(2023)Fang, Wang, Fang, Gou, Liu, Yan, Liu, Xie, and Lu]{fang2023anygrasp}
Hao-Shu Fang, Chenxi Wang, Hongjie Fang, Minghao Gou, Jirong Liu, Hengxu Yan, Wenhai Liu, Yichen Xie, and Cewu Lu.
\newblock Anygrasp: Robust and efficient grasp perception in spatial and temporal domains.
\newblock \emph{IEEE Transactions on Robotics}, 2023.

\bibitem[Finn et~al.(2017)Finn, Yu, Zhang, Abbeel, and Levine]{Finn2017OneShotVI}
Chelsea Finn, Tianhe Yu, T. Zhang, P. Abbeel, and Sergey Levine.
\newblock One-shot visual imitation learning via meta-learning.
\newblock In \emph{CoRL}, 2017.

\bibitem[Florence et~al.(2018)Florence, Manuelli, and Tedrake]{florence2018dense}
Peter~R Florence, Lucas Manuelli, and Russ Tedrake.
\newblock Dense object nets: Learning dense visual object descriptors by and for robotic manipulation.
\newblock \emph{arXiv preprint arXiv:1806.08756}, 2018.

\bibitem[Fu et~al.(2024)Fu, Zhao, Wu, Wetzstein, and Finn]{fu2024humanplus}
Zipeng Fu, Qingqing Zhao, Qi Wu, Gordon Wetzstein, and Chelsea Finn.
\newblock Humanplus: Humanoid shadowing and imitation from humans.
\newblock \emph{arXiv preprint arXiv:2406.10454}, 2024.

\bibitem[Gao et~al.(2024)Gao, Zhang, Xu, Cai, and Shao]{gao2024flip}
Chongkai Gao, Haozhuo Zhang, Zhixuan Xu, Zhehao Cai, and Lin Shao.
\newblock Flip: Flow-centric generative planning for general-purpose manipulation tasks.
\newblock \emph{arXiv preprint arXiv:2412.08261}, 2024.

\bibitem[Gao et~al.(2025)Gao, Zhou, Du, Zhang, and Gan]{gao2025adaworld}
Shenyuan Gao, Siyuan Zhou, Yilun Du, Jun Zhang, and Chuang Gan.
\newblock Adaworld: Learning adaptable world models with latent actions.
\newblock \emph{arXiv preprint arXiv:2503.18938}, 2025.

\bibitem[Gervet et~al.(2023)Gervet, Chintala, Batra, Malik, and Chaplot]{gervet2023navigating}
Theophile Gervet, Soumith Chintala, Dhruv Batra, Jitendra Malik, and Devendra~Singh Chaplot.
\newblock Navigating to objects in the real world.
\newblock \emph{Science Robotics}, 2023.

\bibitem[Gleicher(1998)]{gleicher1998retargetting}
Michael Gleicher.
\newblock Retargetting motion to new characters.
\newblock In \emph{Proceedings of the 25th annual conference on Computer graphics and interactive techniques}, pages 33--42, 1998.

\bibitem[Guo et~al.(2025)Guo, Huo, Shi, Song, Zhang, and Zhao]{guo2025t2vphysbench}
Xuyang Guo, Jiayan Huo, Zhenmei Shi, Zhao Song, Jiahao Zhang, and Jiale Zhao.
\newblock T2vphysbench: A first-principles benchmark for physical consistency in text-to-video generation.
\newblock \emph{arXiv preprint arXiv:2505.00337}, 2025.

\bibitem[Hartley and Zisserman(2003)]{hartley2003multiple}
Richard Hartley and Andrew Zisserman.
\newblock \emph{Multiple view geometry in computer vision}.
\newblock Cambridge university press, 2003.

\bibitem[He et~al.(2024)He, Luo, Xiao, Zhang, Kitani, Liu, and Shi]{he2024learning}
Tairan He, Zhengyi Luo, Wenli Xiao, Chong Zhang, Kris Kitani, Changliu Liu, and Guanya Shi.
\newblock Learning human-to-humanoid real-time whole-body teleoperation.
\newblock In \emph{2024 IEEE/RSJ International Conference on Intelligent Robots and Systems (IROS)}, pages 8944--8951. IEEE, 2024.

\bibitem[He et~al.(2022{\natexlab{a}})He, Sun, Wang, Huang, Bao, and Zhou]{he2022onepose++}
Xingyi He, Jiaming Sun, Yuang Wang, Di Huang, Hujun Bao, and Xiaowei Zhou.
\newblock Onepose++: Keypoint-free one-shot object pose estimation without cad models.
\newblock \emph{Advances in Neural Information Processing Systems}, 35:\penalty0 35103--35115, 2022{\natexlab{a}}.

\bibitem[He et~al.(2020)He, Sun, Huang, Liu, Fan, and Sun]{he2020pvn3d}
Yisheng He, Wei Sun, Haibin Huang, Jianran Liu, Haoqiang Fan, and Jian Sun.
\newblock Pvn3d: A deep point-wise 3d keypoints voting network for 6dof pose estimation.
\newblock In \emph{Proceedings of the IEEE/CVF conference on computer vision and pattern recognition}, pages 11632--11641, 2020.

\bibitem[He et~al.(2021)He, Huang, Fan, Chen, and Sun]{he2021ffb6d}
Yisheng He, Haibin Huang, Haoqiang Fan, Qifeng Chen, and Jian Sun.
\newblock Ffb6d: A full flow bidirectional fusion network for 6d pose estimation.
\newblock In \emph{Proceedings of the IEEE/CVF conference on computer vision and pattern recognition}, pages 3003--3013, 2021.

\bibitem[He et~al.(2022{\natexlab{b}})He, Wang, Fan, Sun, and Chen]{he2022fs6d}
Yisheng He, Yao Wang, Haoqiang Fan, Jian Sun, and Qifeng Chen.
\newblock Fs6d: Few-shot 6d pose estimation of novel objects.
\newblock In \emph{Proceedings of the IEEE/CVF Conference on Computer Vision and Pattern Recognition}, pages 6814--6824, 2022{\natexlab{b}}.

\bibitem[Hsu et~al.(2024)Hsu, Wen, Xu, Narang, Wang, Zhu, Biswas, and Birchfield]{hsu2024spot}
Cheng-Chun Hsu, Bowen Wen, Jie Xu, Yashraj Narang, Xiaolong Wang, Yuke Zhu, Joydeep Biswas, and Stan Birchfield.
\newblock Spot: Se (3) pose trajectory diffusion for object-centric manipulation.
\newblock \emph{arXiv preprint arXiv:2411.00965}, 2024.

\bibitem[Hu et~al.(2014)Hu, Ott, and Lee]{hu2014online}
Kai Hu, Christian Ott, and Dongheui Lee.
\newblock Online human walking imitation in task and joint space based on quadratic programming.
\newblock In \emph{2014 IEEE International Conference on Robotics and Automation (ICRA)}, pages 3458--3464. IEEE, 2014.

\bibitem[Huang et~al.(2023)Huang, Wang, Zhang, Li, Wu, and Fei-Fei]{huang2023voxposer}
Wenlong Huang, Chen Wang, Ruohan Zhang, Yunzhu Li, Jiajun Wu, and Li Fei-Fei.
\newblock Voxposer: Composable 3d value maps for robotic manipulation with language models.
\newblock \emph{arXiv preprint arXiv:2307.05973}, 2023.

\bibitem[Huang et~al.(2024{\natexlab{a}})Huang, Wang, Li, Zhang, and Fei-Fei]{huang2024rekep}
Wenlong Huang, Chen Wang, Yunzhu Li, Ruohan Zhang, and Li Fei-Fei.
\newblock Rekep: Spatio-temporal reasoning of relational keypoint constraints for robotic manipulation.
\newblock \emph{arXiv preprint arXiv:2409.01652}, 2024{\natexlab{a}}.

\bibitem[Huang et~al.(2024{\natexlab{b}})Huang, Zhang, Xu, He, Yu, Dong, Ma, Chanpaisit, Si, Jiang, Wang, Chen, Chen, Wang, Lin, Qiao, and Liu]{huang2024vbench++}
Ziqi Huang, Fan Zhang, Xiaojie Xu, Yinan He, Jiashuo Yu, Ziyue Dong, Qianli Ma, Nattapol Chanpaisit, Chenyang Si, Yuming Jiang, Yaohui Wang, Xinyuan Chen, Ying-Cong Chen, Limin Wang, Dahua Lin, Yu Qiao, and Ziwei Liu.
\newblock Vbench++: Comprehensive and versatile benchmark suite for video generative models.
\newblock \emph{arXiv preprint arXiv:2411.13503}, 2024{\natexlab{b}}.

\bibitem[Jiang et~al.(2023)Jiang, Chen, Liu, Yu, Yu, and Chen]{jiang2023motiongpt}
Biao Jiang, Xin Chen, Wen Liu, Jingyi Yu, Gang Yu, and Tao Chen.
\newblock Motiongpt: Human motion as a foreign language.
\newblock \emph{Advances in Neural Information Processing Systems}, 36:\penalty0 20067--20079, 2023.

\bibitem[Ju et~al.(2024)Ju, Hu, Zhang, Zhang, Jiang, and Xu]{ju2024robo}
Yuanchen Ju, Kaizhe Hu, Guowei Zhang, Gu Zhang, Mingrun Jiang, and Huazhe Xu.
\newblock Robo-abc: Affordance generalization beyond categories via semantic correspondence for robot manipulation.
\newblock In \emph{European Conference on Computer Vision}, pages 222--239. Springer, 2024.

\bibitem[Karaev et~al.(2024)Karaev, Makarov, Wang, Neverova, Vedaldi, and Rupprecht]{karaev2024cotracker3}
Nikita Karaev, Iurii Makarov, Jianyuan Wang, Natalia Neverova, Andrea Vedaldi, and Christian Rupprecht.
\newblock Cotracker3: Simpler and better point tracking by pseudo-labelling real videos.
\newblock \emph{arXiv preprint arXiv:2410.11831}, 2024.

\bibitem[Karamcheti et~al.(2023)Karamcheti, Nair, Chen, Kollar, Finn, Sadigh, and Liang]{karamcheti2023language}
Siddharth Karamcheti, Suraj Nair, Annie~S Chen, Thomas Kollar, Chelsea Finn, Dorsa Sadigh, and Percy Liang.
\newblock Language-driven representation learning for robotics.
\newblock \emph{arXiv preprint arXiv:2302.12766}, 2023.

\bibitem[Kareer et~al.(2024)Kareer, Patel, Punamiya, Mathur, Cheng, Wang, Hoffman, and Xu]{kareer2024egomimic}
Simar Kareer, Dhruv Patel, Ryan Punamiya, Pranay Mathur, Shuo Cheng, Chen Wang, Judy Hoffman, and Danfei Xu.
\newblock Egomimic: Scaling imitation learning via egocentric video.
\newblock \emph{arXiv preprint arXiv:2410.24221}, 2024.

\bibitem[Ke et~al.(2024)Ke, Narnhofer, Huang, Ke, Peters, Fragkiadaki, Obukhov, and Schindler]{ke2024rollingdepth}
Bingxin Ke, Dominik Narnhofer, Shengyu Huang, Lei Ke, Torben Peters, Katerina Fragkiadaki, Anton Obukhov, and Konrad Schindler.
\newblock Video depth without video models, 2024.

\bibitem[Kerbl et~al.(2023)Kerbl, Kopanas, Leimk{\"u}hler, and Drettakis]{kerbl20233d}
Bernhard Kerbl, Georgios Kopanas, Thomas Leimk{\"u}hler, and George Drettakis.
\newblock 3d gaussian splatting for real-time radiance field rendering.
\newblock \emph{ACM Trans. Graph.}, 42\penalty0 (4):\penalty0 139--1, 2023.

\bibitem[Kerr et~al.(2024)Kerr, Kim, Wu, Yi, Wang, Goldberg, and Kanazawa]{kerr2024robot}
Justin Kerr, Chung~Min Kim, Mingxuan Wu, Brent Yi, Qianqian Wang, Ken Goldberg, and Angjoo Kanazawa.
\newblock Robot see robot do: Imitating articulated object manipulation with monocular 4d reconstruction.
\newblock \emph{arXiv preprint arXiv:2409.18121}, 2024.

\bibitem[Kim et~al.(2024)Kim, Wu, Kerr, Goldberg, Tancik, and Kanazawa]{kim2024garfield}
Chung~Min Kim, Mingxuan Wu, Justin Kerr, Ken Goldberg, Matthew Tancik, and Angjoo Kanazawa.
\newblock Garfield: Group anything with radiance fields.
\newblock In \emph{Proceedings of the IEEE/CVF Conference on Computer Vision and Pattern Recognition}, pages 21530--21539, 2024.

\bibitem[Ko et~al.(2023)Ko, Mao, Du, Sun, and Tenenbaum]{ko2023learning}
Po-Chen Ko, Jiayuan Mao, Yilun Du, Shao-Hua Sun, and Joshua~B Tenenbaum.
\newblock Learning to act from actionless videos through dense correspondences.
\newblock \emph{arXiv preprint arXiv:2310.08576}, 2023.

\bibitem[Kuindersma et~al.(2016)Kuindersma, Deits, Fallon, Valenzuela, Dai, Permenter, Koolen, Marion, and Tedrake]{kuindersma2016optimization}
Scott Kuindersma, Robin Deits, Maurice Fallon, Andr{\'e}s Valenzuela, Hongkai Dai, Frank Permenter, Twan Koolen, Pat Marion, and Russ Tedrake.
\newblock Optimization-based locomotion planning, estimation, and control design for the atlas humanoid robot.
\newblock \emph{Autonomous robots}, 40:\penalty0 429--455, 2016.

\bibitem[Labb{\'e} et~al.(2020)Labb{\'e}, Carpentier, Aubry, and Sivic]{labbe2020cosypose}
Yann Labb{\'e}, Justin Carpentier, Mathieu Aubry, and Josef Sivic.
\newblock Cosypose: Consistent multi-view multi-object 6d pose estimation.
\newblock In \emph{Computer Vision--ECCV 2020: 16th European Conference, Glasgow, UK, August 23--28, 2020, Proceedings, Part XVII 16}, pages 574--591. Springer, 2020.

\bibitem[Labb\'e et~al.(2022)Labb\'e, Manuelli, Mousavian, Tyree, Birchfield, Tremblay, Carpentier, Aubry, Fox, and Sivic]{labbe2022megapose}
Yann Labb\'e, Lucas Manuelli, Arsalan Mousavian, Stephen Tyree, Stan Birchfield, Jonathan Tremblay, Justin Carpentier, Mathieu Aubry, Dieter Fox, and Josef Sivic.
\newblock Megapose: 6d pose estimation of novel objects via render \& compare.
\newblock In \emph{Proceedings of the 6th Conference on Robot Learning (CoRL)}, 2022.

\bibitem[Lakshmipathy et~al.(2024)Lakshmipathy, Hodgins, and Pollard]{lakshmipathy2024kinematic}
Arjun~S Lakshmipathy, Jessica~K Hodgins, and Nancy~S Pollard.
\newblock Kinematic motion retargeting for contact-rich anthropomorphic manipulations.
\newblock \emph{arXiv preprint arXiv:2402.04820}, 2024.

\bibitem[Lepert et~al.(2025)Lepert, Fang, and Bohg]{lepert2025phantom}
Marion Lepert, Jiaying Fang, and Jeannette Bohg.
\newblock Phantom: Training robots without robots using only human videos.
\newblock \emph{arXiv preprint arXiv:2503.00779}, 2025.

\bibitem[Lepetit et~al.(2009)Lepetit, Moreno-Noguer, and Fua]{lepetit2009ep}
Vincent Lepetit, Francesc Moreno-Noguer, and Pascal Fua.
\newblock Ep n p: An accurate o (n) solution to the p n p problem.
\newblock \emph{International journal of computer vision}, 81:\penalty0 155--166, 2009.

\bibitem[Li et~al.(2023{\natexlab{a}})Li, Vutukur, Yu, Shugurov, Busam, Yang, and Ilic]{li2023nerf}
Fu Li, Shishir~Reddy Vutukur, Hao Yu, Ivan Shugurov, Benjamin Busam, Shaowu Yang, and Slobodan Ilic.
\newblock Nerf-pose: A first-reconstruct-then-regress approach for weakly-supervised 6d object pose estimation.
\newblock In \emph{Proceedings of the IEEE/CVF International Conference on Computer Vision}, pages 2123--2133, 2023{\natexlab{a}}.

\bibitem[Li et~al.(2024{\natexlab{a}})Li, Sun, Sevilla-Lara, and Jampani]{li2024one}
Gen Li, Deqing Sun, Laura Sevilla-Lara, and Varun Jampani.
\newblock One-shot open affordance learning with foundation models.
\newblock In \emph{Proceedings of the IEEE/CVF Conference on Computer Vision and Pattern Recognition}, pages 3086--3096, 2024{\natexlab{a}}.

\bibitem[Li et~al.(2024{\natexlab{b}})Li, Tsagkas, Song, Mon-Williams, Vijayakumar, Shao, and Sevilla-Lara]{li2024learning}
Gen Li, Nikolaos Tsagkas, Jifei Song, Ruaridh Mon-Williams, Sethu Vijayakumar, Kun Shao, and Laura Sevilla-Lara.
\newblock Learning precise affordances from egocentric videos for robotic manipulation.
\newblock \emph{arXiv preprint arXiv:2408.10123}, 2024{\natexlab{b}}.

\bibitem[Li et~al.(2024{\natexlab{c}})Li, Zhu, Xie, Jiang, Seo, Pavlakos, and Zhu]{li2024okami}
Jinhan Li, Yifeng Zhu, Yuqi Xie, Zhenyu Jiang, Mingyo Seo, Georgios Pavlakos, and Yuke Zhu.
\newblock Okami: Teaching humanoid robots manipulation skills through single video imitation.
\newblock In \emph{8th Annual Conference on Robot Learning}, 2024{\natexlab{c}}.

\bibitem[Li et~al.(2023{\natexlab{b}})Li, Zhu, Han, Hou, Guo, and Cheng]{licvpr23amt}
Zhen Li, Zuo-Liang Zhu, Ling-Hao Han, Qibin Hou, Chun-Le Guo, and Ming-Ming Cheng.
\newblock Amt: All-pairs multi-field transforms for efficient frame interpolation.
\newblock In \emph{IEEE Conference on Computer Vision and Pattern Recognition (CVPR)}, 2023{\natexlab{b}}.

\bibitem[Liang et~al.(2024)Liang, Liu, Ozguroglu, Sudhakar, Dave, Tokmakov, Song, and Vondrick]{liang2024dreamitate}
Junbang Liang, Ruoshi Liu, Ege Ozguroglu, Sruthi Sudhakar, Achal Dave, Pavel Tokmakov, Shuran Song, and Carl Vondrick.
\newblock Dreamitate: Real-world visuomotor policy learning via video generation.
\newblock \emph{arXiv preprint arXiv:2406.16862}, 2024.

\bibitem[Liang et~al.(2021)Liang, Li, Wang, Xiong, Mao, and Zhang]{liang2021dynamic}
Yuwei Liang, Weijie Li, Yue Wang, Rong Xiong, Yichao Mao, and Jiafan Zhang.
\newblock Dynamic movement primitive based motion retargeting for dual-arm sign language motions.
\newblock In \emph{2021 IEEE International Conference on Robotics and Automation (ICRA)}, pages 8195--8201. IEEE, 2021.

\bibitem[Liu et~al.(2024{\natexlab{a}})Liu, Sun, Wang, Wang, Sun, Ye, Zhang, and Duan]{liu2024reconx}
Fangfu Liu, Wenqiang Sun, Hanyang Wang, Yikai Wang, Haowen Sun, Junliang Ye, Jun Zhang, and Yueqi Duan.
\newblock Reconx: Reconstruct any scene from sparse views with video diffusion model.
\newblock \emph{arXiv preprint arXiv:2408.16767}, 2024{\natexlab{a}}.

\bibitem[Liu et~al.(2023)Liu, Zeng, Ren, Li, Zhang, Yang, Li, Yang, Su, Zhu, et~al.]{liu2023grounding}
Shilong Liu, Zhaoyang Zeng, Tianhe Ren, Feng Li, Hao Zhang, Jie Yang, Chunyuan Li, Jianwei Yang, Hang Su, Jun Zhu, et~al.
\newblock Grounding dino: Marrying dino with grounded pre-training for open-set object detection.
\newblock \emph{arXiv preprint arXiv:2303.05499}, 2023.

\bibitem[Liu et~al.(2024{\natexlab{b}})Liu, Wang, Zhang, Zhang, Tombari, and Ji]{liu2024unopose}
Xingyu Liu, Gu Wang, Ruida Zhang, Chenyangguang Zhang, Federico Tombari, and Xiangyang Ji.
\newblock Unopose: Unseen object pose estimation with an unposed rgb-d reference image.
\newblock \emph{arXiv preprint arXiv:2411.16106}, 2024{\natexlab{b}}.

\bibitem[Liu et~al.(2018)Liu, Gupta, Abbeel, and Levine]{liu2018imitation}
YuXuan Liu, Abhishek Gupta, Pieter Abbeel, and Sergey Levine.
\newblock Imitation from observation: Learning to imitate behaviors from raw video via context translation.
\newblock In \emph{2018 IEEE International Conference on Robotics and Automation (ICRA)}, pages 1118--1125. IEEE, 2018.

\bibitem[Liu et~al.(2022)Liu, Wen, Peng, Lin, Long, Komura, and Wang]{liu2022gen6d}
Yuan Liu, Yilin Wen, Sida Peng, Cheng Lin, Xiaoxiao Long, Taku Komura, and Wenping Wang.
\newblock Gen6d: Generalizable model-free 6-dof object pose estimation from rgb images.
\newblock In \emph{European Conference on Computer Vision}, pages 298--315. Springer, 2022.

\bibitem[Luo et~al.(2023)Luo, Cao, Kitani, Xu, et~al.]{luo2023perpetual}
Zhengyi Luo, Jinkun Cao, Kris Kitani, Weipeng Xu, et~al.
\newblock Perpetual humanoid control for real-time simulated avatars.
\newblock In \emph{Proceedings of the IEEE/CVF International Conference on Computer Vision}, pages 10895--10904, 2023.

\bibitem[Mandlekar et~al.(2018)Mandlekar, Zhu, Garg, Booher, Spero, Tung, Gao, Emmons, Gupta, Orbay, et~al.]{mandlekar2018roboturk}
Ajay Mandlekar, Yuke Zhu, Animesh Garg, Jonathan Booher, Max Spero, Albert Tung, Julian Gao, John Emmons, Anchit Gupta, Emre Orbay, et~al.
\newblock Roboturk: A crowdsourcing platform for robotic skill learning through imitation.
\newblock In \emph{Conference on Robot Learning}, pages 879--893. PMLR, 2018.

\bibitem[Mendonca et~al.(2023)Mendonca, Bahl, and Pathak]{mendonca2023structured}
Russell Mendonca, Shikhar Bahl, and Deepak Pathak.
\newblock Structured world models from human videos.
\newblock \emph{arXiv preprint arXiv:2308.10901}, 2023.

\bibitem[Motamed et~al.(2025)Motamed, Culp, Swersky, Jaini, and Geirhos]{motamed2025generative}
Saman Motamed, Laura Culp, Kevin Swersky, Priyank Jaini, and Robert Geirhos.
\newblock Do generative video models learn physical principles from watching videos?
\newblock \emph{arXiv preprint arXiv:2501.09038}, 2025.

\bibitem[Nakaoka et~al.(2005)Nakaoka, Nakazawa, Kanehiro, Kaneko, Morisawa, and Ikeuchi]{nakaoka2005task}
Shinichiro Nakaoka, Atsushi Nakazawa, Fumio Kanehiro, Kenji Kaneko, Mitsuharu Morisawa, and Katsushi Ikeuchi.
\newblock Task model of lower body motion for a biped humanoid robot to imitate human dances.
\newblock In \emph{2005 IEEE/RSJ International Conference on Intelligent Robots and Systems}, pages 3157--3162. IEEE, 2005.

\bibitem[Nguyen et~al.(2024)Nguyen, Groueix, Salzmann, and Lepetit]{nguyen2024gigapose}
Van~Nguyen Nguyen, Thibault Groueix, Mathieu Salzmann, and Vincent Lepetit.
\newblock Gigapose: Fast and robust novel object pose estimation via one correspondence.
\newblock In \emph{Proceedings of the IEEE/CVF Conference on Computer Vision and Pattern Recognition}, pages 9903--9913, 2024.

\bibitem[Niekum et~al.(2012)Niekum, Osentoski, Konidaris, and Barto]{niekum2012learning}
Scott Niekum, Sarah Osentoski, George Konidaris, and Andrew~G Barto.
\newblock Learning and generalization of complex tasks from unstructured demonstrations.
\newblock In \emph{2012 IEEE/RSJ International Conference on Intelligent Robots and Systems}, pages 5239--5246. IEEE, 2012.

\bibitem[O'Neill et~al.(2023)O'Neill, Rehman, Gupta, Maddukuri, Gupta, Padalkar, Lee, Pooley, Gupta, Mandlekar, et~al.]{o2023open}
Abby O'Neill, Abdul Rehman, Abhinav Gupta, Abhiram Maddukuri, Abhishek Gupta, Abhishek Padalkar, Abraham Lee, Acorn Pooley, Agrim Gupta, Ajay Mandlekar, et~al.
\newblock Open x-embodiment: Robotic learning datasets and rt-x models.
\newblock \emph{arXiv preprint arXiv:2310.08864}, 2023.

\bibitem[{\"O}rnek et~al.(2024){\"O}rnek, Labb\'e, Tekin, Ma, Keskin, Forster, and Hoda{\v{n}}]{ornek2024foundpose}
Evin~P{\i}nar {\"O}rnek, Yann Labb\'e, Bugra Tekin, Lingni Ma, Cem Keskin, Christian Forster, and Tom{\'a}{\v{s}} Hoda{\v{n}}.
\newblock Foundpose: Unseen object pose estimation with foundation features.
\newblock \emph{European Conference on Computer Vision (ECCV)}, 2024.

\bibitem[Park et~al.(2019)Park, Patten, and Vincze]{park2019pix2pose}
Kiru Park, Timothy Patten, and Markus Vincze.
\newblock Pix2pose: Pixel-wise coordinate regression of objects for 6d pose estimation.
\newblock In \emph{Proceedings of the IEEE/CVF international conference on computer vision}, pages 7668--7677, 2019.

\bibitem[Park et~al.(2020)Park, Mousavian, Xiang, and Fox]{park2020latentfusion}
Keunhong Park, Arsalan Mousavian, Yu Xiang, and Dieter Fox.
\newblock Latentfusion: End-to-end differentiable reconstruction and rendering for unseen object pose estimation.
\newblock In \emph{Proceedings of the IEEE/CVF conference on computer vision and pattern recognition}, pages 10710--10719, 2020.

\bibitem[Patel et~al.(2022)Patel, Wang, Radosavovic, and Malik]{patel2022learning}
Austin Patel, Andrew Wang, Ilija Radosavovic, and Jitendra Malik.
\newblock Learning to imitate object interactions from internet videos.
\newblock \emph{arXiv preprint arXiv:2211.13225}, 2022.

\bibitem[Patel et~al.(2023)Patel, Eghbalzadeh, Kamra, Iuzzolino, Jain, and Desai]{vlamp}
Dhruvesh Patel, Hamid Eghbalzadeh, Nitin Kamra, Michael~Louis Iuzzolino, Unnat Jain, and Ruta Desai.
\newblock Pretrained language models as visual planners for human assistance.
\newblock In \emph{ICCV}, 2023.

\bibitem[Patel et~al.(2025)Patel, Yin, Huang, Garg, Nayyeri, Fei-Fei, Lazebnik, and Li]{patel2025real}
Shivansh Patel, Xinchen Yin, Wenlong Huang, Shubham Garg, Hooshang Nayyeri, Li Fei-Fei, Svetlana Lazebnik, and Yunzhu Li.
\newblock A real-to-sim-to-real approach to robotic manipulation with vlm-generated iterative keypoint rewards.
\newblock \emph{arXiv preprint arXiv:2502.08643}, 2025.

\bibitem[Pathak et~al.(2018)Pathak, Mahmoudieh, Luo, Agrawal, Chen, Shentu, Shelhamer, Malik, Efros, and Darrell]{pathak2018zero}
Deepak Pathak, Parsa Mahmoudieh, Guanghao Luo, Pulkit Agrawal, Dian Chen, Yide Shentu, Evan Shelhamer, Jitendra Malik, Alexei~A Efros, and Trevor Darrell.
\newblock Zero-shot visual imitation.
\newblock In \emph{Proceedings of the IEEE conference on computer vision and pattern recognition workshops}, pages 2050--2053, 2018.

\bibitem[Penco et~al.(2019)Penco, Scianca, Modugno, Lanari, Oriolo, and Ivaldi]{penco2019multimode}
Luigi Penco, Nicola Scianca, Valerio Modugno, Leonardo Lanari, Giuseppe Oriolo, and Serena Ivaldi.
\newblock A multimode teleoperation framework for humanoid loco-manipulation: An application for the icub robot.
\newblock \emph{IEEE Robotics \& Automation Magazine}, 26\penalty0 (4):\penalty0 73--82, 2019.

\bibitem[Peng et~al.(2021)Peng, Ma, Abbeel, Levine, and Kanazawa]{peng2021amp}
Xue~Bin Peng, Ze Ma, Pieter Abbeel, Sergey Levine, and Angjoo Kanazawa.
\newblock Amp: Adversarial motion priors for stylized physics-based character control.
\newblock \emph{ACM Transactions on Graphics (ToG)}, 40\penalty0 (4):\penalty0 1--20, 2021.

\bibitem[Ponimatkin et~al.(2025)Ponimatkin, C{\'\i}fka, Sou{\v{c}}ek, Fourmy, Labb{\'e}, Petrik, and Sivic]{ponimatkin20256d}
Georgy Ponimatkin, Martin C{\'\i}fka, Tom{\'a}{\v{s}} Sou{\v{c}}ek, M{\'e}d{\'e}ric Fourmy, Yann Labb{\'e}, Vladimir Petrik, and Josef Sivic.
\newblock 6d object pose tracking in internet videos for robotic manipulation.
\newblock \emph{arXiv preprint arXiv:2503.10307}, 2025.

\bibitem[Qin et~al.(2022)Qin, Wu, Liu, Jiang, Yang, Fu, and Wang]{qin2022dexmv}
Yuzhe Qin, Yueh-Hua Wu, Shaowei Liu, Hanwen Jiang, Ruihan Yang, Yang Fu, and Xiaolong Wang.
\newblock Dexmv: Imitation learning for dexterous manipulation from human videos.
\newblock In \emph{European Conference on Computer Vision}, pages 570--587. Springer, 2022.

\bibitem[Radford et~al.(2021)Radford, Kim, Hallacy, Ramesh, Goh, Agarwal, Sastry, Askell, Mishkin, Clark, Krueger, and Sutskever]{radford2021learningtransferablevisualmodels}
Alec Radford, Jong~Wook Kim, Chris Hallacy, Aditya Ramesh, Gabriel Goh, Sandhini Agarwal, Girish Sastry, Amanda Askell, Pamela Mishkin, Jack Clark, Gretchen Krueger, and Ilya Sutskever.
\newblock Learning transferable visual models from natural language supervision, 2021.

\bibitem[Ravi et~al.(2024)Ravi, Gabeur, Hu, Hu, Ryali, Ma, Khedr, Rädle, Rolland, Gustafson, Mintun, Pan, Alwala, Carion, Wu, Girshick, Dollár, and Feichtenhofer]{ravi2024sam2segmentimages}
Nikhila Ravi, Valentin Gabeur, Yuan-Ting Hu, Ronghang Hu, Chaitanya Ryali, Tengyu Ma, Haitham Khedr, Roman Rädle, Chloe Rolland, Laura Gustafson, Eric Mintun, Junting Pan, Kalyan~Vasudev Alwala, Nicolas Carion, Chao-Yuan Wu, Ross Girshick, Piotr Dollár, and Christoph Feichtenhofer.
\newblock Sam 2: Segment anything in images and videos, 2024.

\bibitem[Ren et~al.(2025)Ren, Sundaresan, Sadigh, Choudhury, and Bohg]{ren2025motion}
Juntao Ren, Priya Sundaresan, Dorsa Sadigh, Sanjiban Choudhury, and Jeannette Bohg.
\newblock Motion tracks: A unified representation for human-robot transfer in few-shot imitation learning.
\newblock \emph{arXiv preprint arXiv:2501.06994}, 2025.

\bibitem[Sermanet et~al.(2018)Sermanet, Lynch, Chebotar, Hsu, Jang, Schaal, Levine, and Brain]{sermanet2018time}
Pierre Sermanet, Corey Lynch, Yevgen Chebotar, Jasmine Hsu, Eric Jang, Stefan Schaal, Sergey Levine, and Google Brain.
\newblock Time-contrastive networks: Self-supervised learning from video.
\newblock In \emph{2018 IEEE international conference on robotics and automation (ICRA)}, pages 1134--1141. IEEE, 2018.

\bibitem[Sharma et~al.(2018)Sharma, Mohan, Pinto, and Gupta]{sharma2018multiple}
Pratyusha Sharma, Lekha Mohan, Lerrel Pinto, and Abhinav Gupta.
\newblock Multiple interactions made easy (mime): Large scale demonstrations data for imitation.
\newblock \emph{arXiv:1810.07121}, 2018.

\bibitem[Sharma et~al.(2019)Sharma, Pathak, and Gupta]{sharma2019third}
Pratyusha Sharma, Deepak Pathak, and Abhinav Gupta.
\newblock Third-person visual imitation learning via decoupled hierarchical controller.
\newblock \emph{Advances in Neural Information Processing Systems}, 32, 2019.

\bibitem[Shi et~al.(2025)Shi, Zhao, Wang, Pedroza, Luo, Wang, Ma, and Jayaraman]{shi2025zeromimic}
Junyao Shi, Zhuolun Zhao, Tianyou Wang, Ian Pedroza, Amy Luo, Jie Wang, Jason Ma, and Dinesh Jayaraman.
\newblock Zeromimic: Distilling robotic manipulation skills from web videos.
\newblock \emph{arXiv preprint arXiv:2503.23877}, 2025.

\bibitem[Shugurov et~al.(2022)Shugurov, Li, Busam, and Ilic]{shugurov2022osop}
Ivan Shugurov, Fu Li, Benjamin Busam, and Slobodan Ilic.
\newblock Osop: A multi-stage one shot object pose estimation framework.
\newblock In \emph{Proceedings of the IEEE/CVF Conference on Computer Vision and Pattern Recognition}, pages 6835--6844, 2022.

\bibitem[Sivakumar et~al.(2022)Sivakumar, Shaw, and Pathak]{sivakumar2022robotic}
Aravind Sivakumar, Kenneth Shaw, and Deepak Pathak.
\newblock Robotic telekinesis: Learning a robotic hand imitator by watching humans on youtube.
\newblock \emph{arXiv preprint arXiv:2202.10448}, 2022.

\bibitem[Smith et~al.(2019)Smith, Dhawan, Zhang, Abbeel, and Levine]{smith2019avid}
Laura Smith, Nikita Dhawan, Marvin Zhang, Pieter Abbeel, and Sergey Levine.
\newblock Avid: Learning multi-stage tasks via pixel-level translation of human videos.
\newblock \emph{arXiv preprint arXiv:1912.04443}, 2019.

\bibitem[Srirama et~al.(2024)Srirama, Dasari, Bahl, and Gupta]{srirama2024hrp}
Mohan~Kumar Srirama, Sudeep Dasari, Shikhar Bahl, and Abhinav Gupta.
\newblock Hrp: Human affordances for robotic pre-training.
\newblock \emph{arXiv preprint arXiv:2407.18911}, 2024.

\bibitem[Sun et~al.(2022)Sun, Wang, Zhang, He, Zhao, Zhang, and Zhou]{sun2022onepose}
Jiaming Sun, Zihao Wang, Siyu Zhang, Xingyi He, Hongcheng Zhao, Guofeng Zhang, and Xiaowei Zhou.
\newblock Onepose: One-shot object pose estimation without cad models.
\newblock In \emph{Proceedings of the IEEE/CVF Conference on Computer Vision and Pattern Recognition}, pages 6825--6834, 2022.

\bibitem[Sun et~al.(2024)Sun, Zhou, Yuan, Sun, Li, Jia, Adam, Hariharan, Zhao, and Liu]{sun2024video}
Yihong Sun, Hao Zhou, Liangzhe Yuan, Jennifer~J Sun, Yandong Li, Xuhui Jia, Hartwig Adam, Bharath Hariharan, Long Zhao, and Ting Liu.
\newblock Video creation by demonstration.
\newblock \emph{arXiv preprint arXiv:2412.09551}, 2024.

\bibitem[Tremblay et~al.(2018)Tremblay, To, Sundaralingam, Xiang, Fox, and Birchfield]{tremblay2018deep}
Jonathan Tremblay, Thang To, Balakumar Sundaralingam, Yu Xiang, Dieter Fox, and Stan Birchfield.
\newblock Deep object pose estimation for semantic robotic grasping of household objects.
\newblock \emph{arXiv preprint arXiv:1809.10790}, 2018.

\bibitem[Valassakis et~al.(2022)Valassakis, Papagiannis, Di~Palo, and Johns]{valassakis2022demonstrate}
Eugene Valassakis, Georgios Papagiannis, Norman Di~Palo, and Edward Johns.
\newblock Demonstrate once, imitate immediately (dome): Learning visual servoing for one-shot imitation learning.
\newblock In \emph{2022 IEEE/RSJ International Conference on Intelligent Robots and Systems (IROS)}, pages 8614--8621. IEEE, 2022.

\bibitem[Vecerik et~al.(2024)Vecerik, Doersch, Yang, Davchev, Aytar, Zhou, Hadsell, Agapito, and Scholz]{vecerik2024robotap}
Mel Vecerik, Carl Doersch, Yi Yang, Todor Davchev, Yusuf Aytar, Guangyao Zhou, Raia Hadsell, Lourdes Agapito, and Jon Scholz.
\newblock Robotap: Tracking arbitrary points for few-shot visual imitation.
\newblock In \emph{2024 IEEE International Conference on Robotics and Automation (ICRA)}, pages 5397--5403. IEEE, 2024.

\bibitem[Wang et~al.(2023{\natexlab{a}})Wang, Fan, Sun, Zhang, Fei-Fei, Xu, Zhu, and Anandkumar]{wang2023mimicplay}
Chen Wang, Linxi Fan, Jiankai Sun, Ruohan Zhang, Li Fei-Fei, Danfei Xu, Yuke Zhu, and Anima Anandkumar.
\newblock Mimicplay: Long-horizon imitation learning by watching human play.
\newblock \emph{arXiv preprint arXiv:2302.12422}, 2023{\natexlab{a}}.

\bibitem[Wang et~al.(2024)Wang, Liu, Guo, Zhou, and Atkeson]{wang2024one}
Jianren Wang, Kangni Liu, Dingkun Guo, Xian Zhou, and Christopher~G Atkeson.
\newblock One-shot video imitation via parameterized symbolic abstraction graphs.
\newblock \emph{arXiv preprint arXiv:2408.12674}, 2024.

\bibitem[Wang et~al.(2023{\natexlab{b}})Wang, He, Li, Li, Yu, Ma, Li, Chen, Chen, Wang, et~al.]{wang2023internvid}
Yi Wang, Yinan He, Yizhuo Li, Kunchang Li, Jiashuo Yu, Xin Ma, Xinhao Li, Guo Chen, Xinyuan Chen, Yaohui Wang, et~al.
\newblock Internvid: A large-scale video-text dataset for multimodal understanding and generation.
\newblock In \emph{The Twelfth International Conference on Learning Representations}, 2023{\natexlab{b}}.

\bibitem[Wasserman et~al.(2023)Wasserman, Yadav, Chowdhary, Gupta, and Jain]{sling}
Justin Wasserman, Karmesh Yadav, Girish Chowdhary, Abhinav Gupta, and Unnat Jain.
\newblock Last-mile embodied visual navigation.
\newblock In \emph{Conference on Robot Learning}, pages 666--678. PMLR, 2023.

\bibitem[Wasserman et~al.(2024)Wasserman, Chowdhary, Gupta, and Jain]{xgx}
Justin Wasserman, Girish Chowdhary, Abhinav Gupta, and Unnat Jain.
\newblock Exploitation-guided exploration for semantic embodied navigation.
\newblock In \emph{2024 IEEE International Conference on Robotics and Automation (ICRA)}, pages 2901--2908. IEEE, 2024.

\bibitem[Wen et~al.(2023{\natexlab{a}})Wen, Tremblay, Blukis, Tyree, Muller, Evans, Fox, Kautz, and Birchfield]{wen2023bundlesdf}
Bowen Wen, Jonathan Tremblay, Valts Blukis, Stephen Tyree, Thomas Muller, Alex Evans, Dieter Fox, Jan Kautz, and Stan Birchfield.
\newblock Bundlesdf: Neural 6-dof tracking and 3d reconstruction of unknown objects.
\newblock \emph{CVPR}, 2023{\natexlab{a}}.

\bibitem[Wen et~al.(2023{\natexlab{b}})Wen, Yang, Kautz, and Birchfield]{wen2023foundationpose}
Bowen Wen, Wei Yang, Jan Kautz, and Stan Birchfield.
\newblock Foundationpose: Unified 6d pose estimation and tracking of novel objects.
\newblock \emph{arXiv preprint arXiv:2312.08344}, 2023{\natexlab{b}}.

\bibitem[Wu et~al.(2024)Wu, Wang, Chen, Eppner, and Liu]{wu2024one}
Albert Wu, Ruocheng Wang, Sirui Chen, Clemens Eppner, and C~Karen Liu.
\newblock One-shot transfer of long-horizon extrinsic manipulation through contact retargeting.
\newblock In \emph{2024 IEEE/RSJ International Conference on Intelligent Robots and Systems (IROS)}, pages 13891--13898. IEEE, 2024.

\bibitem[Xie et~al.(2024)Xie, Lee, Xiao, and Finn]{xie2024decomposing}
Annie Xie, Lisa Lee, Ted Xiao, and Chelsea Finn.
\newblock Decomposing the generalization gap in imitation learning for visual robotic manipulation.
\newblock In \emph{2024 IEEE International Conference on Robotics and Automation (ICRA)}, pages 3153--3160. IEEE, 2024.

\bibitem[Xu et~al.(2022)Xu, Zhang, Cai, Rezatofighi, and Tao]{xu2022gmflow}
Haofei Xu, Jing Zhang, Jianfei Cai, Hamid Rezatofighi, and Dacheng Tao.
\newblock Gmflow: Learning optical flow via global matching.
\newblock In \emph{Proceedings of the IEEE/CVF conference on computer vision and pattern recognition}, pages 8121--8130, 2022.

\bibitem[Xu et~al.(2023)Xu, Xu, Chi, Veloso, and Song]{xu2023xskill}
Mengda Xu, Zhenjia Xu, Cheng Chi, Manuela Veloso, and Shuran Song.
\newblock Xskill: Cross embodiment skill discovery.
\newblock In \emph{Conference on robot learning}, pages 3536--3555. PMLR, 2023.

\bibitem[Xu et~al.(2024)Xu, Xu, Xu, Chi, Wetzstein, Veloso, and Song]{xu2024flow}
Mengda Xu, Zhenjia Xu, Yinghao Xu, Cheng Chi, Gordon Wetzstein, Manuela Veloso, and Shuran Song.
\newblock Flow as the cross-domain manipulation interface.
\newblock \emph{arXiv preprint arXiv:2407.15208}, 2024.

\bibitem[Yang et~al.(2023)Yang, Du, Ghasemipour, Tompson, Schuurmans, and Abbeel]{yang2023learning}
Mengjiao Yang, Yilun Du, Kamyar Ghasemipour, Jonathan Tompson, Dale Schuurmans, and Pieter Abbeel.
\newblock Learning interactive real-world simulators.
\newblock \emph{arXiv preprint arXiv:2310.06114}, 1\penalty0 (2):\penalty0 6, 2023.

\bibitem[Yang et~al.(2025)Yang, Li, Zhang, Yin, Bai, Ma, Wang, Cai, Wong, Lu, et~al.]{yang2025vlipp}
Xindi Yang, Baolu Li, Yiming Zhang, Zhenfei Yin, Lei Bai, Liqian Ma, Zhiyong Wang, Jianfei Cai, Tien-Tsin Wong, Huchuan Lu, et~al.
\newblock Vlipp: Towards physically plausible video generation with vision and language informed physical prior.
\newblock \emph{arXiv e-prints}, pages arXiv--2503, 2025.

\bibitem[Ye et~al.(2024)Ye, Jang, Jeon, Joo, Yang, Peng, Mandlekar, Tan, Chao, Lin, et~al.]{ye2024latent}
Seonghyeon Ye, Joel Jang, Byeongguk Jeon, Sejune Joo, Jianwei Yang, Baolin Peng, Ajay Mandlekar, Reuben Tan, Yu-Wei Chao, Bill~Yuchen Lin, et~al.
\newblock Latent action pretraining from videos.
\newblock \emph{arXiv preprint arXiv:2410.11758}, 2024.

\bibitem[Yu et~al.(2018)Yu, Finn, Xie, Dasari, Zhang, Abbeel, and Levine]{yu2018one}
Tianhe Yu, Chelsea Finn, Annie Xie, Sudeep Dasari, Tianhao Zhang, Pieter Abbeel, and Sergey Levine.
\newblock One-shot imitation from observing humans via domain-adaptive meta-learning.
\newblock \emph{arXiv preprint arXiv:1802.01557}, 2018.

\bibitem[Yuan et~al.(2024)Yuan, Wen, Zhang, and Gao]{yuan2024general}
Chengbo Yuan, Chuan Wen, Tong Zhang, and Yang Gao.
\newblock General flow as foundation affordance for scalable robot learning.
\newblock \emph{arXiv preprint arXiv:2401.11439}, 2024.

\bibitem[Zakka et~al.(2022)Zakka, Zeng, Florence, Tompson, Bohg, and Dwibedi]{zakka2022xirl}
Kevin Zakka, Andy Zeng, Pete Florence, Jonathan Tompson, Jeannette Bohg, and Debidatta Dwibedi.
\newblock Xirl: Cross-embodiment inverse reinforcement learning.
\newblock In \emph{Conference on Robot Learning}, pages 537--546. PMLR, 2022.

\bibitem[Zhang et~al.(2024)Zhang, Zhai, Bautista, Miao, Toshev, Susskind, and Gu]{zhang2024world}
Qihang Zhang, Shuangfei Zhai, Miguel~Angel Bautista, Kevin Miao, Alexander Toshev, Joshua Susskind, and Jiatao Gu.
\newblock World-consistent video diffusion with explicit 3d modeling.
\newblock \emph{arXiv preprint arXiv:2412.01821}, 2024.

\bibitem[Zhang(2000)]{zhang2000flexible}
Zhengyou Zhang.
\newblock A flexible new technique for camera calibration.
\newblock \emph{IEEE Transactions on pattern analysis and machine intelligence}, 22\penalty0 (11):\penalty0 1330--1334, 2000.

\bibitem[Zhao et~al.(2023)Zhao, Kumar, Levine, and Finn]{zhao2023learning}
Tony~Z Zhao, Vikash Kumar, Sergey Levine, and Chelsea Finn.
\newblock Learning fine-grained bimanual manipulation with low-cost hardware.
\newblock \emph{arXiv preprint arXiv:2304.13705}, 2023.

\bibitem[Zhen et~al.(2025)Zhen, Sun, Zhang, Li, Zhou, Du, and Gan]{zhen2025tesseract}
Haoyu Zhen, Qiao Sun, Hongxin Zhang, Junyan Li, Siyuan Zhou, Yilun Du, and Chuang Gan.
\newblock Tesseract: Learning 4d embodied world models.
\newblock 2025.

\bibitem[Zhou et~al.(2025)Zhou, Wang, Tai, Deng, Liu, and Jia]{zhou2025you}
Huayi Zhou, Ruixiang Wang, Yunxin Tai, Yueci Deng, Guiliang Liu, and Kui Jia.
\newblock You only teach once: Learn one-shot bimanual robotic manipulation from video demonstrations.
\newblock \emph{arXiv preprint arXiv:2501.14208}, 2025.

\bibitem[Zhou et~al.(2024)Zhou, Ma, Lin, Wang, Qiu, and Liang]{zhou2024mitigating}
Jiaming Zhou, Teli Ma, Kun-Yu Lin, Zifan Wang, Ronghe Qiu, and Junwei Liang.
\newblock Mitigating the human-robot domain discrepancy in visual pre-training for robotic manipulation.
\newblock \emph{arXiv preprint arXiv:2406.14235}, 2024.

\bibitem[Zhu et~al.(2024)Zhu, Ju, Zhang, Wang, Yuan, Hu, and Xu]{zhu2024densematcher}
Junzhe Zhu, Yuanchen Ju, Junyi Zhang, Muhan Wang, Zhecheng Yuan, Kaizhe Hu, and Huazhe Xu.
\newblock Densematcher: Learning 3d semantic correspondence for category-level manipulation from a single demo.
\newblock \emph{arXiv preprint arXiv:2412.05268}, 2024.

\end{thebibliography}
